\documentclass[runningheads]{llncs}

\usepackage{eccv}

\usepackage{eccvabbrv}

\usepackage{graphicx}
\usepackage{booktabs}
\usepackage{multirow}
\usepackage{arydshln}
\usepackage[accsupp]{axessibility}  

\usepackage{hyperref}

\usepackage{orcidlink}

\begin{document}

\makeatletter
\renewcommand*{\@fnsymbol}[1]{\ensuremath{\ifcase#1\or \dagger\or \ddagger\or
   \mathsection\or \mathparagraph\or \|\or **\or \dagger\dagger
   \or \ddagger\ddagger \else\@ctrerr\fi}}
\makeatother
\newcommand*\samethanks[1][\value{footnote}]{\footnotemark[#1]}
\title{Online Continuous Generalized Category Discovery} 
\titlerunning{Online Continuous Generalized Category Discovery}


\author{Keon-Hee Park\inst{1}\orcidlink{0009-0008-3654-812X} \and
Hakyung Lee\inst{2} \orcidlink{0009-0001-7600-9760} \and
Kyungwoo Song\inst{2}\thanks{Corresponding authors} \orcidlink{0000-0003-0082-4280} \and
Gyeong-Moon~Park\inst{1}\samethanks[1]\orcidlink{0000-0003-4011-9981}}

\authorrunning{K.-H. Park et al.}

\institute{Kyung Hee University, Yongin, Republic of Korea \\
\email{\{khpark,gmpark\}@khu.ac.kr} \and
Yonsei University, Seoul, Republic of Korea \\
\email{\{hakyunglee0417, kyungwoo.song\}@yonsei.ac.kr}}

\maketitle
\begin{abstract}
  With the advancement of deep neural networks in computer vision, artificial intelligence (AI) is widely employed in real-world applications. However, AI still faces limitations in mimicking high-level human capabilities, such as novel category discovery, for practical use. While some methods utilizing offline continual learning have been proposed for novel category discovery, they neglect the continuity of data streams in real-world settings. In this work, we introduce \textit{Online Continuous Generalized Category Discovery} (\textbf{OCGCD}), which considers the dynamic nature of data streams where data can be created and deleted in real time. Additionally, we propose a novel method, \textbf{DEAN}, \textbf{D}iscovery via \textbf{E}nergy guidance and feature \textbf{A}ugmentatio\textbf{N}, which can discover novel categories in an online manner through energy-guided discovery and facilitate discriminative learning via energy-based contrastive loss. Furthermore, DEAN effectively pseudo-labels unlabeled data through variance-based feature augmentation. Experimental results demonstrate that our proposed DEAN achieves outstanding performance in proposed OCGCD scenario. The implementation code is available at \url{https://github.com/KHU-AGI/OCGCD}.
  
  \keywords{Online Continual Learning \and Generalized Category Discovery \and Energy-Guided Discovery \and Variance-based Feature Augmentation}
\end{abstract}

\begin{figure}[t]
  \centering
  \begin{subfigure}[t]{0.42\columnwidth}
    \centering
    \includegraphics[width=\columnwidth]{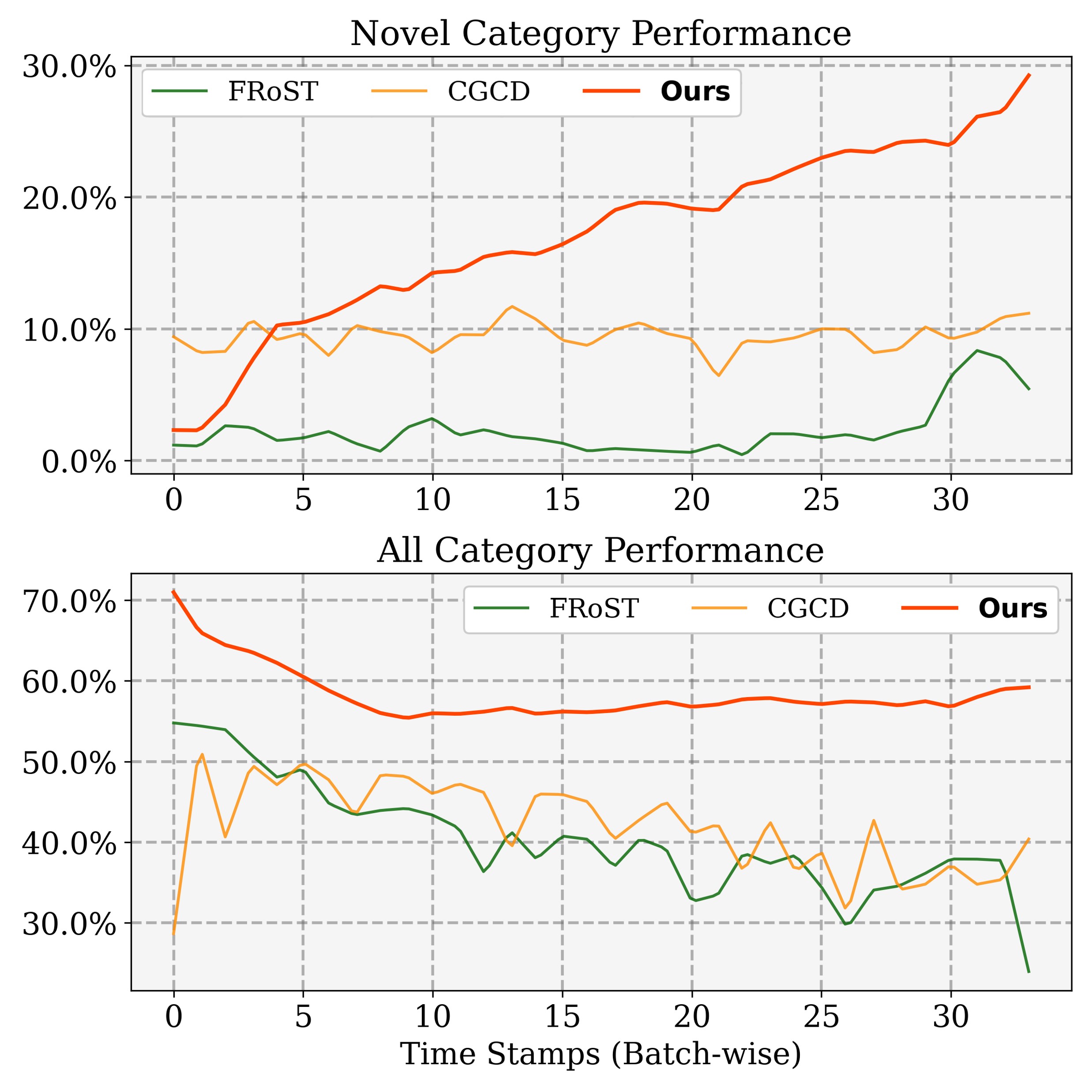}
    \caption{Experiment of prior methods in an online manner on CUB200.}
    \label{fig:motiv_a}
  \end{subfigure}
  \hfill 
  \begin{subfigure}[t]{0.55\columnwidth}
  \centering 
    \includegraphics[width=\columnwidth]{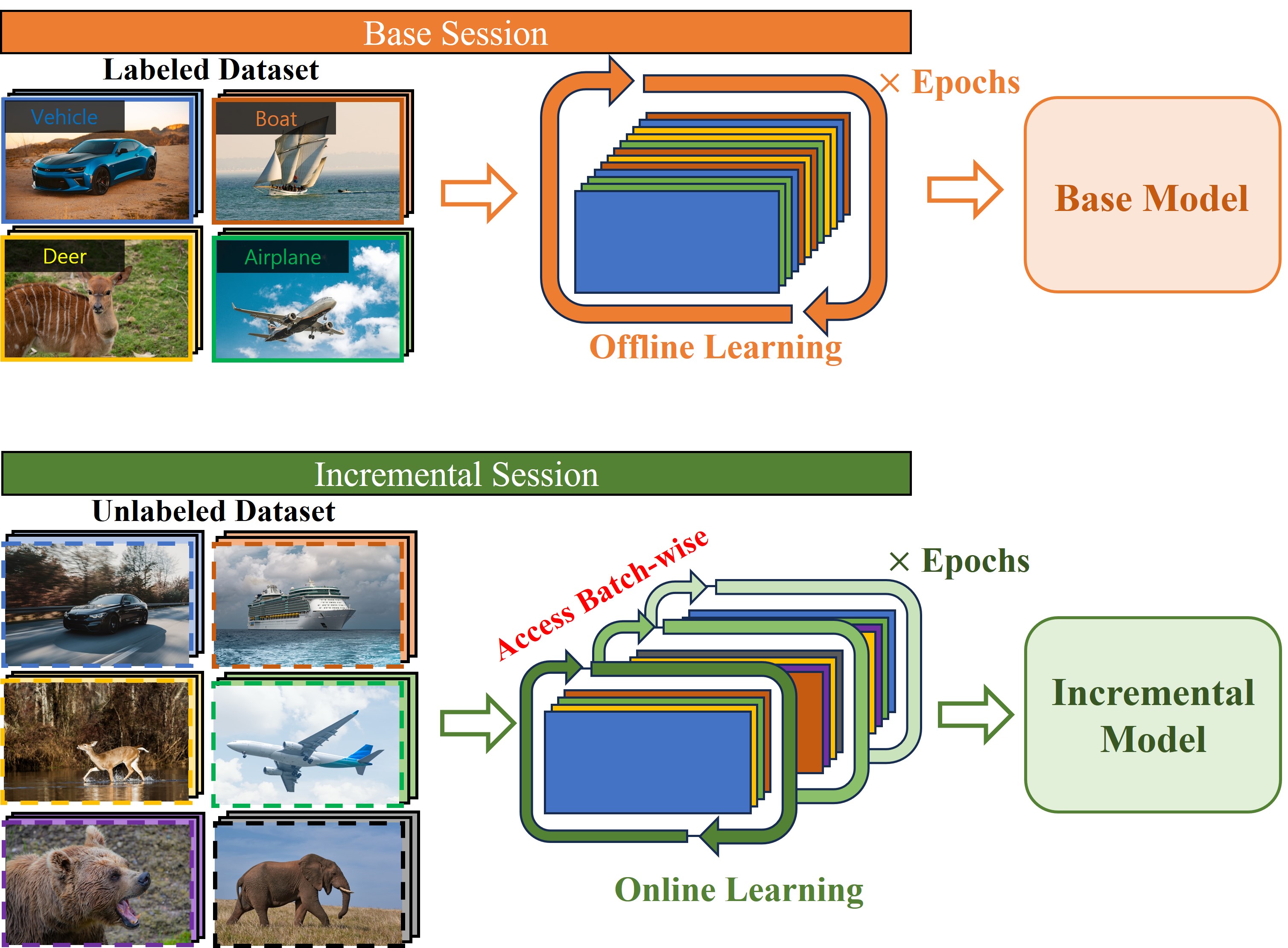}
    \caption{The description of the proposed OCGCD.}
    \label{fig:motiv_scenario}
  \end{subfigure}
  \caption{\Cref{fig:motiv_a} shows that existing methods recorded poor performance in online training, suggesting that prior methods cannot handle online continual learning. \Cref{fig:motiv_scenario} shows the proposed scenario, OCGCD. As our scenario assumes batch-wise online learning, the model suffers severe forgetting and poor novel category discovery.}
  \label{fig:motiv}
\end{figure}

\section{Introduction}
\label{sec:intro}
The rapid progress of deep neural networks in computer vision has greatly facilitated the widespread adoption of artificial intelligence into real-world applications. Nevertheless, the networks still have challenges in embodying high-level human capabilities such as object recognition and grouping. To emulate these sophisticated human abilities, there has been a lot of interest in the field of category discovery. Specifically, generalized category discovery (GCD)~\cite{vaze2022generalized} demands advanced recognition and grouping skills to discover unknown categories within unlabeled datasets.
In general, existing methods for GCD rely on an impractical assumption that they can access both labeled and unlabeled datasets simultaneously during training, which prevents these methods from meeting real-world application demands.

In this context, a new scenario of continuous category discovery \cite{joseph2022novel,roy2022class,zhang2022grow,wu2023metagcd,kim2023proxy} has emerged to address the unrealistic assumption by adopting offline continual learning. Despite its great advantage not requiring the assumption, these offline-based methods have inherent limitations in handling the continuous nature of data streams in the real world.
Current approaches to offline continual learning primarily focus on static data streams, thus failing to adequately accommodate the dynamic nature of data streams where data can be created and deleted in an online manner.
To verify this limitation, we conducted an empirical analysis using the CUB200~\cite{wah2011caltech} dataset, dividing it into 160 labeled categories and 40 unlabeled categories. We evaluated the performance of existing methods ~\cite{roy2022class,kim2023proxy} for continuous category discovery in an online manner. As shown in \Cref{fig:motiv_a}, we observed that existing methods~\cite{roy2022class,kim2023proxy} exhibited inferior performance in identifying novel categories, thereby resulting in poor overall performances. This indicates that the offline-based GCD methods have difficulty in clustering novel data effectively in the online learning scenario.


In light of this observation, in this paper, we introduce a novel online learning scenario in continuous generalized category discovery, called \textbf{Online Continuous Generalized Category Discovery (OCGCD)}. As shown in~\Cref{fig:motiv_scenario}, OCGCD involves batch-wise training of unlabeled data in an online continual manner without any prior knowledge of the unknown data. Therefore, our novel scenario closely replicates real-world data streams characterized by dynamic creation and elimination of data during training~\cite{he2020incremental}.
Here we summarize the key challenges of our novel OCGCD scenario as follows:
1) \textit{Challenging category discovery.} The model struggles to retain the knowledge of trained categories because of the continuous nature of the data, making novel category discovery challenging.
2) \textit{Noisy pseudo-labeling.} Online learning restricts access to the entire training dataset, leading to inaccurate category assignments and the generation of severe noise pseudo-labels for unlabeled data.
3) \textit{Severe catastrophic forgetting.} Online learning erases observed data after training, resulting in severe forgetting of previously acquired knowledge~\cite{mccloskey1989catastrophic}. Hence, our novel OCGCD scenario requires addressing the aforementioned challenges to achieve successful learning outcomes.

To address the challenges of OCGCD, we propose a novel approach called \textbf{DEAN}, \textbf{D}iscovery via \textbf{E}nergy Guidance and Feature \textbf{A}ugmentatio\textbf{N}, which 
utilizes the energy score~\cite{liu2020energy} to discover unknown data and further discriminate whether the data is seen or unseen and employs feature augmentation for better pseudo-labeling. By comparing the energy values of known and unknown data, DEAN effectively identifies unknown data among them, facilitating the clustering of novel categories. On top of that, we introduce an energy-based contrastive loss to enhance learning from unknown samples, promoting the acquisition of more discriminative knowledge during online learning. In addition, we propose a novel variance-based feature augmentation, a simple yet effective approach for accurate novel clustering and pseudo-labeling. In our scenario, since the model can access only batch-wise training data at once, it struggles to cluster data accurately. Variance-based feature augmentation~(VFA) temporally extends the number of feature vectors through variational information to promote accurate clustering.
To mitigate catastrophic forgetting, we employ parameter-efficient tuning, especially LoRA~\cite{hu2021lora} in our framework while preserving the pre-trained weights. Extensive experiments demonstrate that our proposed method, DEAN, enables the network to discover novel categories more effectively in the OCGCD scenario than existing methods.

We summarize the main contributions of this paper as follows:
\begin{itemize}
\item We introduce a novel Online Continuous Generalized Category Discovery (OCGCD) scenario that effectively reflects the online characteristics of real-world data. To tackle this scenario, we propose a novel method called Discovery via Energy Guidance and Feature AugmentatioN (DEAN).
\item To the best of our knowledge, this is the first work to introduce the energy score for novel category discovery. We propose the energy-based contrastive loss to enhance online learning of discovered unknown data.
\item For effective pseudo-labeling, we propose a new variance-based feature augmentation, called VFA. The proposed VFA enhances the sample clustering, leading to the improvement of pseudo-label quality.
\item Comprehensive experiments demonstrate that our novel method achieves significant performance improvements in the OCGCD scenario compared to state-of-the-art models.
\end{itemize}

\section{Related Work}
\subsection{Novel Category Discovery}
\label{sec:ncd}
Novel Category Discovery (NCD) focuses on the task of identifying and categorizing unlabeled data~\cite{troisemaine2023novel}. To effectively discover and classify data, this field requires advanced object recognition and clustering skills. Most of the NCD methods fall into two training scenarios: disjoint training and joint training. The disjoint training scenario, also known as a two-stage approach, starts by pre-training the model on a labeled dataset, followed by fine-tuning on an unlabeled dataset from a different domain~\cite{hsu2017learning,liu2022residual,hsu2019multi,han2019learning}. 
The joint training scenario adopts a more integrated approach, simultaneously utilizing both labeled and unlabeled datasets during the training process~\cite{han2020automatically,zhong2021neighborhood,zhong2021openmix,jia2021joint,fini2021unified}. While this scenario outperforms disjoint training approach~\cite{liu2022residual}, its dependence on the availability of labeled data makes it less feasible in real-world applications. Accordingly, more research has been directed towards designing scenarios where they mitigate constraints of NCD for better alignment with real-world settings. Generalized Category Discovery (GCD)~\cite{vaze2022generalized} broadens the setting of NCD by allowing test data to contain both known and unknown classes, in contrast to NCD which focuses on exclusively unknown classes. Moreover, to better represent real-world scenarios, several methods~\cite{joseph2022novel,roy2022class,zhang2022grow,wu2023metagcd,kim2023proxy} have been proposed that integrate continual learning into category discovery. However, existing methods predominantly utilizes an offline continual learning framework, which remains an impractical assumption for real-world applications. 

We propose an online version of continuous GCD that tackles the challenges of online learning through energy-guided clustering and energy-based contrastive loss. Energy-guided clustering utilizes the trained classifier to extract energy from each sample, efficiently discovering novel categories. Moreover, applying energy-based contrastive loss enables the model to effectively learn feature knowledge, thereby enhancing the capacity of the model in online continual learning.
\subsection{Online Continual Learning}
\label{sec:ocl}
Online Continual Learning (OCL) is a more realistic continual learning scenario where the model learns from data arriving in small, sequential batches in real-time, without access to previously seen batches~\cite{mai2022online}. A key aspect of continual learning is to mitigate catastrophic forgetting and balance existing knowledge while adapting to new knowledge~\cite{french1999catastrophic,mccloskey1989catastrophic,mcclelland1995there}. The dynamic distribution of online setting intensifies this challenge, and various approaches have been introduced to address this issue.  
MIR~\cite{aljundi2019online} proposed a memory retrieval technique for replay, selecting mostly interfered samples. 
OCM~\cite{guo2022online} deals with forgetting by maximizing mutual information between past and current data to learn more transferable features. 
i-Blurry~\cite{koh2021online} introduces a realistic setting that combines disjoint and blurry class distributions, mitigating forgetting by employing sample-wise importance sampling. Furthermore, Si-Blurry~\cite{moon2023online} proposed a new blurry task scenario, acknowledging the stochastic properties in real-world data distributions.

In this paper, we propose a novel setting that integrates online continual learning with generalized category discovery. By emphasizing the online learning, OCGCD presents a new perspective in the field of category discovery. It is a more realistic and challenging scenario than existing approaches in GCD, taking into account the continuous flow of real-world data.

\subsection{Noise Label}
\label{sec:nl}
Deep neural networks can overfit to noisy label data, leading to a substantial decrease in the robustness and generalization ability~\cite{zhang2021understanding,arpit2017closer}. Consequently, tackling noisy label data becomes essential, especially in category discovery where pseudo-labeling of unlabeled data plays a pivotal role in identifying new categories. 
Recent studies have primarily focused on discriminating between clean data and noisy label data. DivideMix~\cite{li2020dividemix} use a Gaussian Mixture Model (GMM) generated from the training dataset to differentiate between clean and noisy data, while AugDesc~\cite{nishi2021augmentation} applies various data augmentations to the training dataset to amplify the distinction. SplitNet~\cite{kim2022splitnet} introduces a compact learnable module that separates clean and noisy label data.

In OCGCD, learning occurs without prior knowledge of novel categories, relying only on the current batch-wise data to predict the number of clusters and generate pseudo-labels based on this prediction. This scenario makes it difficult to apply existing techniques that depend on utilizing the training dataset or employing learnable modules. Therefore, we propose Variance-based Feature Augmentation (VFA) to effectively separate clean and noisy labels in our framework. VFA can diversely augment (re-generate) the feature vectors of given data, enabling the creation of discriminative clusters for accurate pseudo-labeling. Unlike CGCD~\cite{kim2023proxy}, which requires training additional modules for the discovery, VFA facilitates end-to-end training without the additional modules or extra learning. 

\section{Online Continuous Generalized Category Discovery}
\subsection{Problem Formulation and Method Overview }
\begin{figure}[t]
    \centering
    \includegraphics[width=\columnwidth]{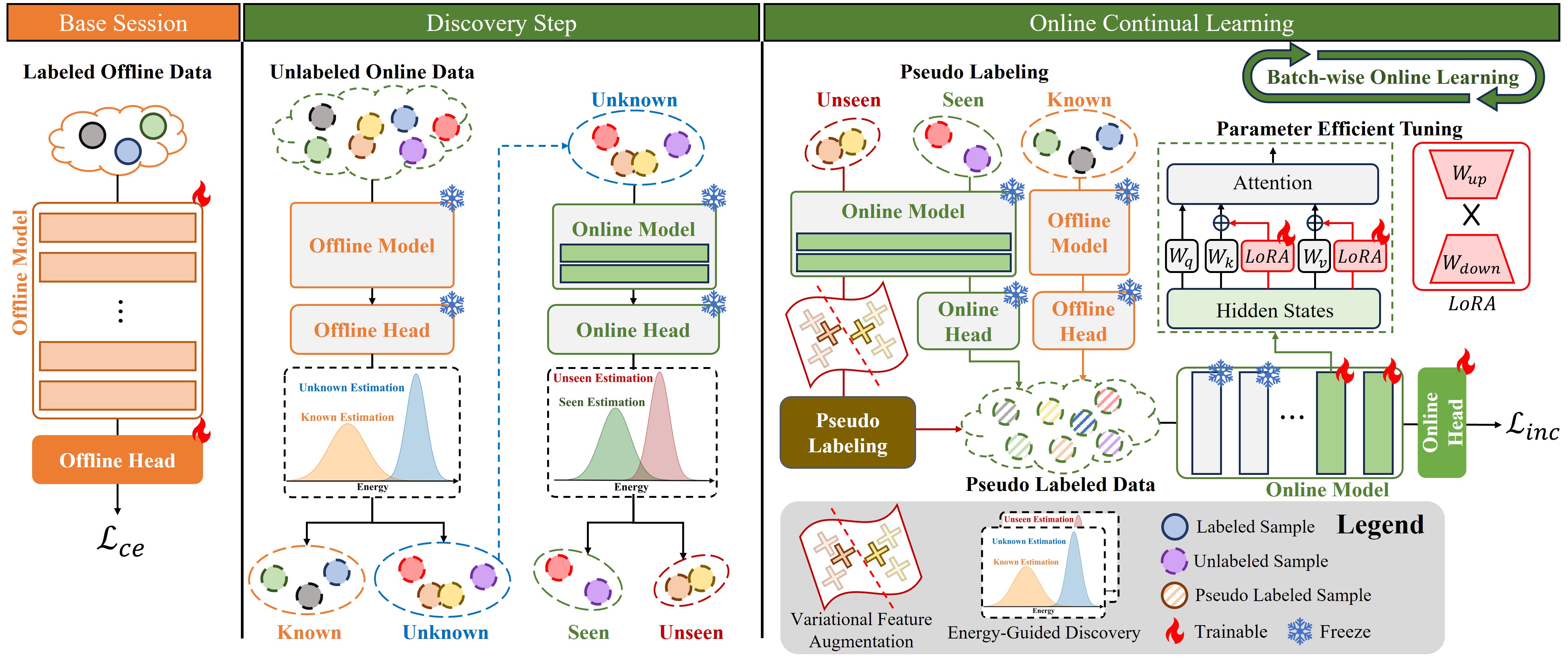}
    \caption{Overall process of the proposed DEAN framework. The energy-guided discovery splits unlabeled data into known, seen, and unseen data for better novel category discovery, while variance-based feature augmentation enhances the clustering of unseen data. $\mathcal{L}_{ec}$ facilitates better discriminative learning in the online continual learning.}
    \label{fig:main}
\end{figure}

\textbf{Problem Formulation.} In online continuous GCD, the training dataset $\mathcal{D}=\{ \mathcal{D}^{base}, \mathcal{D}^{inc}\}$ is given sequentially, where $\mathcal{D}^{base}=\{(x_i,y_i)\in\mathcal{X}\times\mathcal{Y}_l\}_{i=1}^{\left|\mathcal{D}^{base}\right|}$ is the labeled dataset for the base session and $\mathcal{D}^{inc}=\{(x_i,y_i) \in \mathcal{X}\times\mathcal{Y}_u\}_{i=1}^{\left|\mathcal{D}^{inc}\right|}$ is the unlabeled dataset for the incremental session. 
The proposed novel scenario assumes an overlap between the labeled category set $\mathcal{Y}_l$ and the unlabeled category set $\mathcal{Y}_u$, as it is a sub-category of the generalized category discovery.
In this setting, the network trains on $\mathcal{D}^{base}$ and on $\mathcal{D}^{inc}$ in an offline and online manner, respectively.
Our proposed online continuous GCD aims to facilitate the discovery of novel categories in real-world scenarios while preserving the clustering knowledge obtained from the offline training of the labeled dataset.

\noindent
\textbf{Method Overview.} \Cref{fig:main} shows an overview of the proposed method, where the pre-trained vision transformer~\cite{dosovitskiy2020image} via self-supervised learning serves as the backbone network. We initially train the network using the labeled dataset in the base session (\Cref{sec:base_learn}). We introduce energy-guided discovery to identify unknown samples from the given data. Our energy-guided discovery involves a two-step process to determine whether data belong to known or unknown categories and whether the unknown data belong to previously seen or unseen categories (\Cref{sec:energy}). We split the data into known, seen, and unseen categories through energy-guided discovery. For pseudo-labeling of unseen data, we propose variance-based feature augmentation to improve the clustering of unseen data, resulting in better pseudo-labeling (\Cref{sec:vfr}). Subsequently, the network trains pseudo-labeled data in an online manner using an energy-based contrastive loss for better discriminative learning (\Cref{sec:inc_learn}).

\subsection{Supervised Learning at Base Session}
\label{sec:base_learn}
During the base session, we train the offline model $\theta^{\textit{off}}$ consisting of the feature extractor $f^{\textit{off}}(\cdot):{\mathcal{X} \rightarrow \mathbb{R}^{d}}$ and classifier head $g^{\textit{off}}(\cdot): \mathbb{R}^{d} \rightarrow \mathcal{Y}_l$ in an offline manner. As the offline network $\theta^{\textit{off}}=\{f^{\textit{off}}, g^{\textit{off}}\}$ trains on the labeled dataset during the base session, we employ the cross-entropy loss for supervised learning. While existing novel category discovery methods often utilize self-supervised~\cite{gutmann2010noise} and supervised~\cite{khosla2020supervised} contrastive loss for representation learning without a classifier head, our approach utilizes a classifier for clustering. The formulation of the base session training is as follows:
\begin{align}
    \mathcal{L}_{ce}=-\mathbb{E}_{(\mathcal{X},\mathcal{Y}_l)\sim\mathcal{D}^{\textit{base}}}\frac{1}{\left|\mathcal{D}^{\textit{base}}\right|} \sum_{i=1}^{\left|\mathcal{D}^{\textit{base}}\right|}y_{i}\cdot \mathrm{log}\, \delta(g^{\textit{off}}(f^{\textit{off}}(x_i))),
\end{align}
where $\delta(\cdot)$ is the softmax function. We train the model to capture representation knowledge during the base session training. After training the base session, we freeze the offline model and utilize the model at the discovery step to distinguish between known and unknown data, enabling accurate discovery. In addition, we initialize the online model, denoted as $\theta^{\textit{on}}$, based on the parameters of the offline model for the online continual learning on the unlabeled dataset.

\subsection{Energy-Guided Discovery to Identify Unknowns}
\label{sec:energy}
\begin{figure}[t]
  \begin{subfigure}[b]{0.43\columnwidth}
  \includegraphics[width=\linewidth]{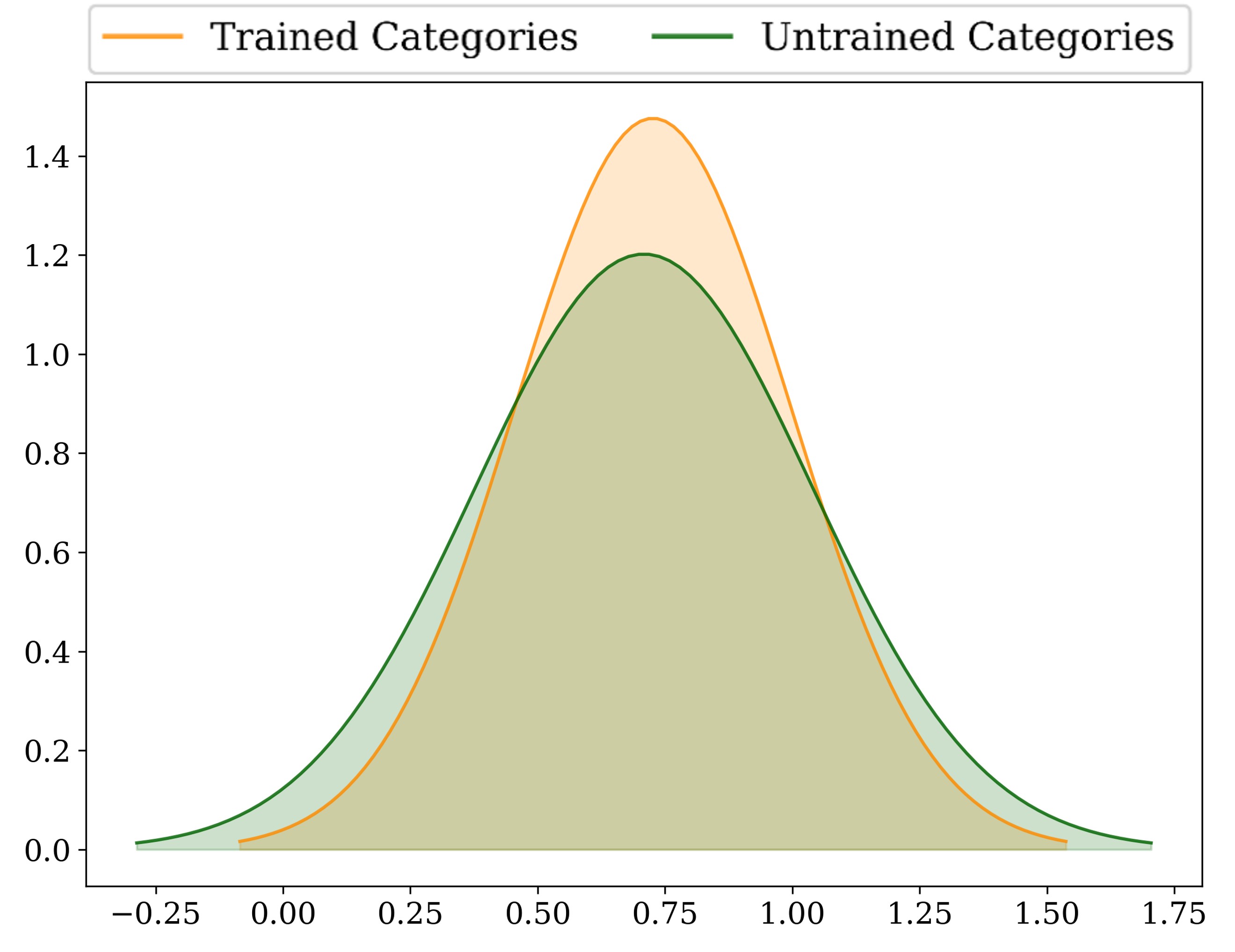}
    \caption{Comparing normal distributions of known and unknown samples on FGVC-Aircraft using fine-split of CGCD.}
    \end{subfigure}
  \hfill 
  \begin{subfigure}[b]{0.43\columnwidth } 
  \centering
  \includegraphics[width=\linewidth]{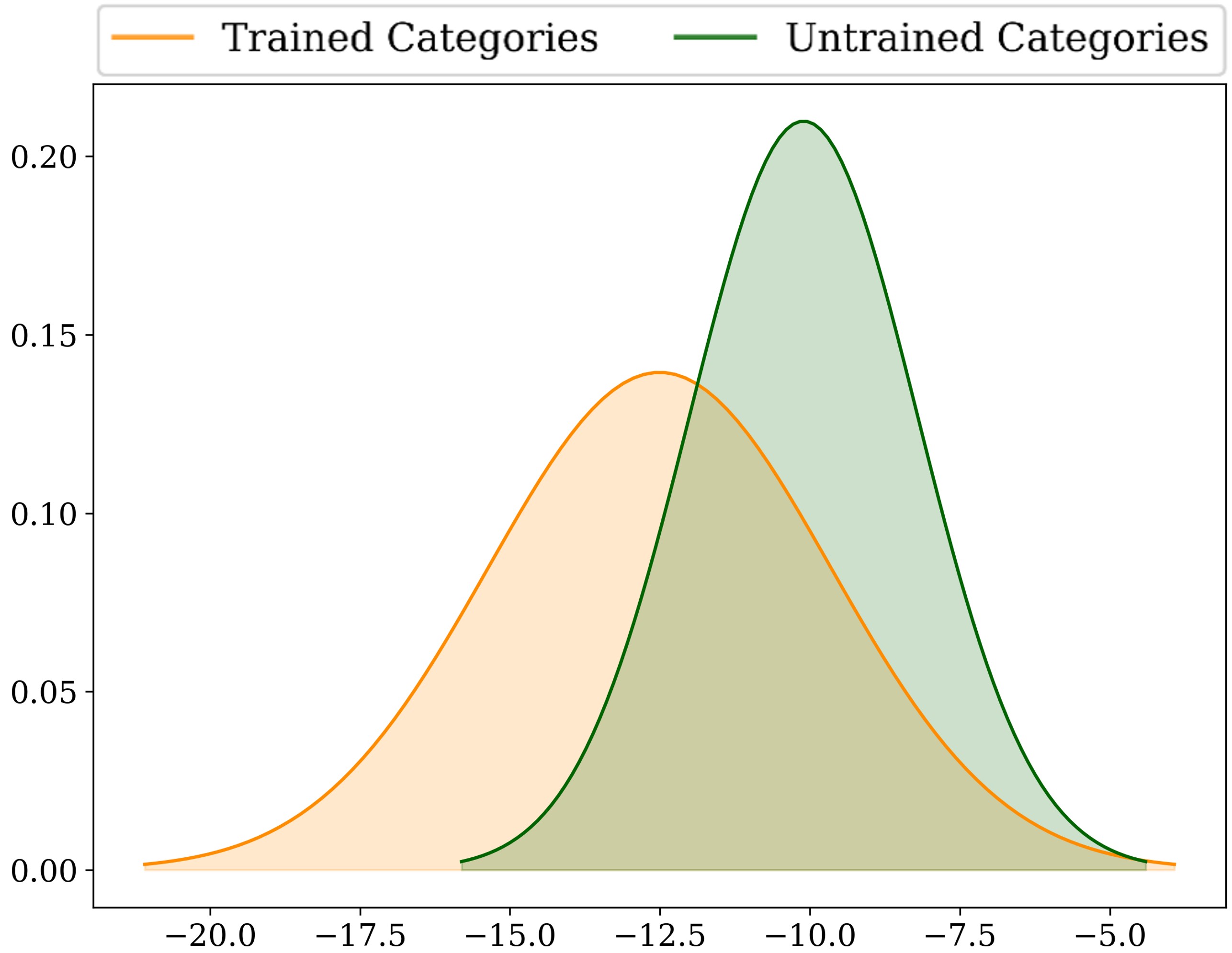}
    \caption{Comparing normal distributions of known and unknown samples on FGVC-Aircraft using \textbf{energy scores}.}
  \end{subfigure} 
\vfill
  \begin{subfigure}[b]{0.43\columnwidth } 
    \centering
    \includegraphics[width=\linewidth]{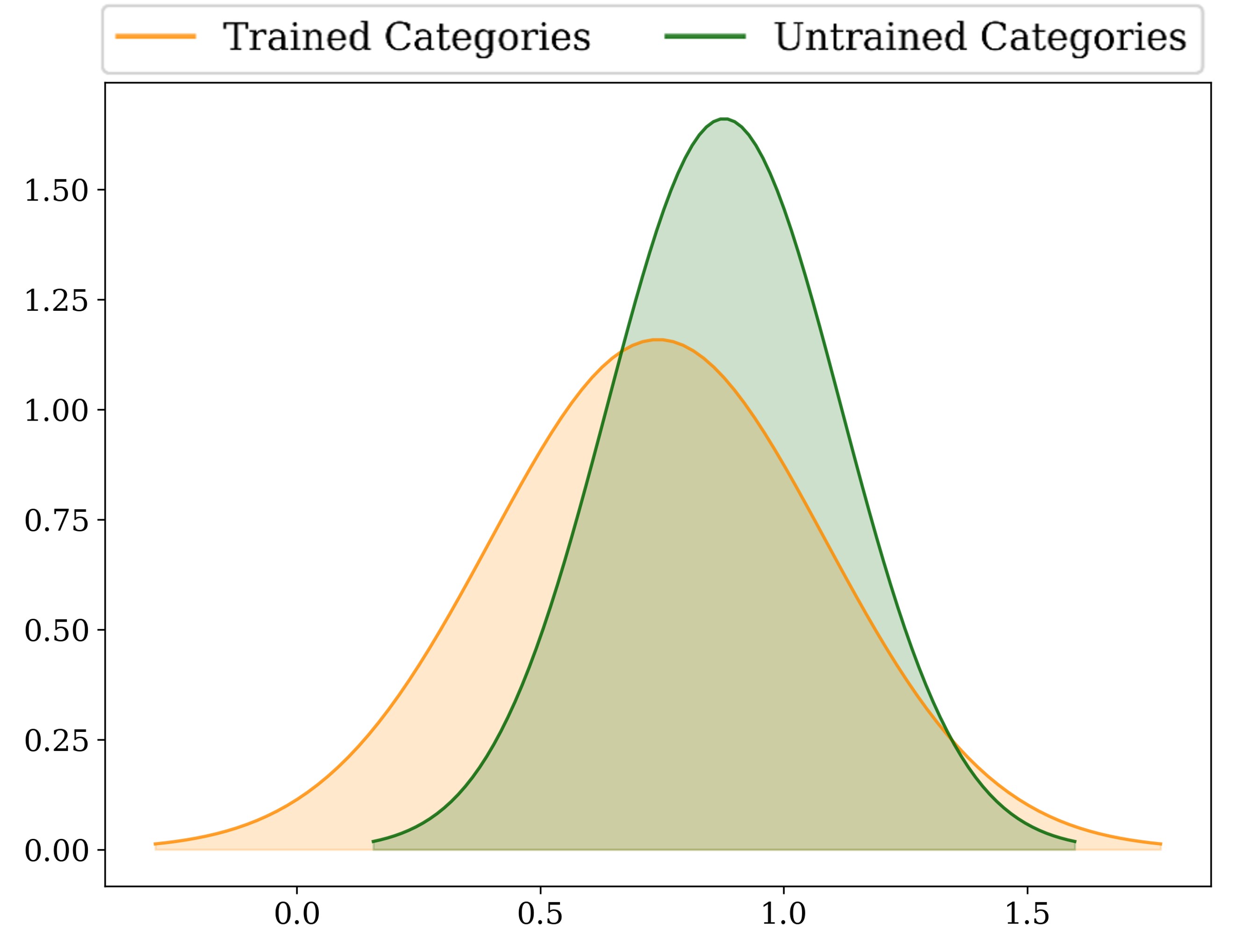}
    \caption{Comparing normal distributions of known and unknown samples on CUB200 using fine-split of CGCD.}
\end{subfigure}
  \hfill
  \begin{subfigure}[b]{0.43\columnwidth } 
  \centering
  \includegraphics[width=\linewidth]{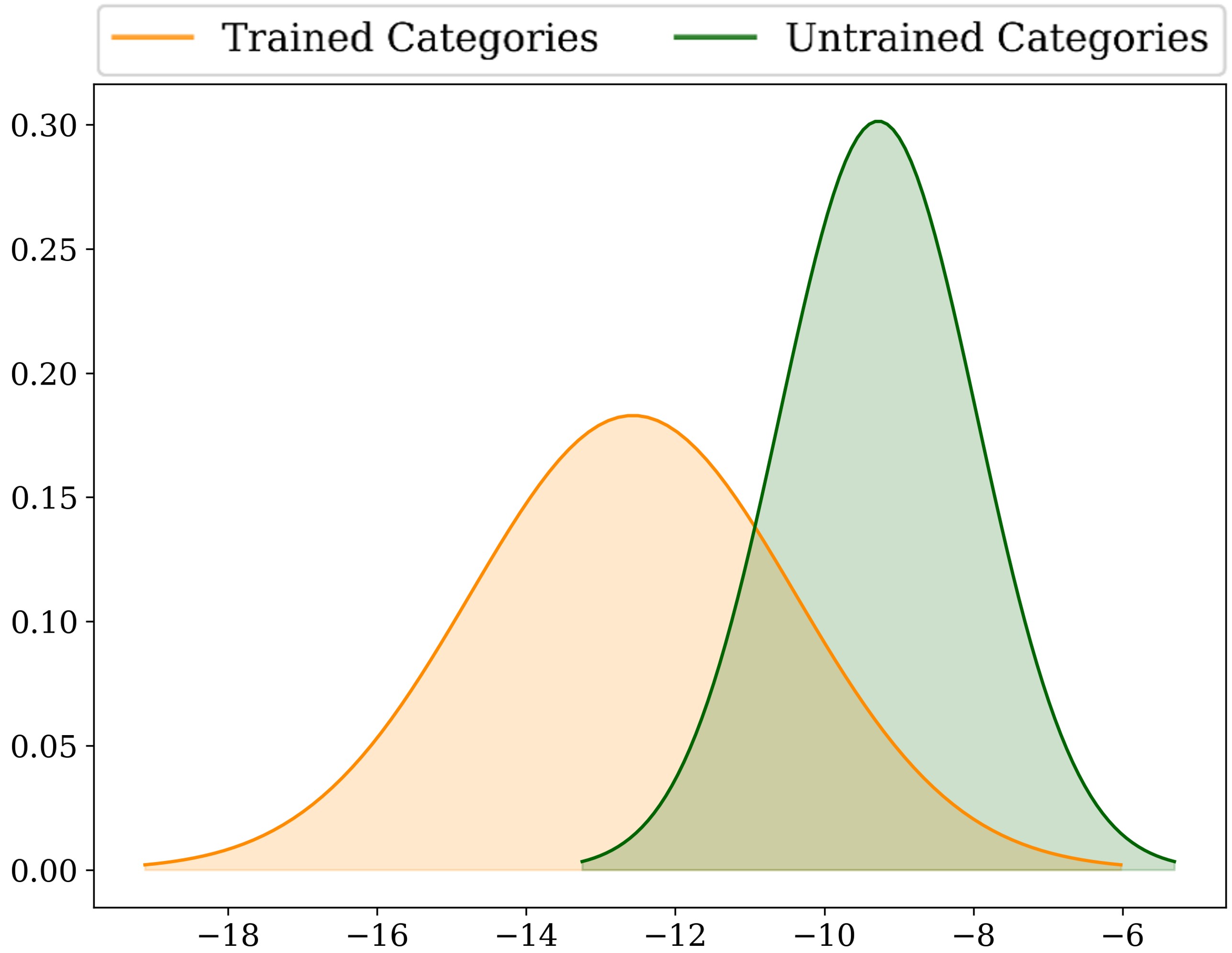}
    \caption{Comparing normal distributions of known and unknown samples on CUB200 using \textbf{energy scores}.}
  \end{subfigure} 
  \caption{Validating the effectiveness of energy scores for novel category discovery without prior knowledge about novel categories by comparison with existing methods.}
    \label{fig:method_energy}
\end{figure}
The discovery step aims to discover new categories within the provided batch-wise data. While CGCD, which assumes the absence of prior knowledge about novel categories, introduced a two-stage split process that depends on learning an additional split network to detect novel data, it has some limitations in our proposed scenario. CGCD includes the requirement for extra parameters and additional learning stage. Moreover, in our proposed online scenario where the model cannot access the entire training data, the split network faces difficulty in identifying novel samples due to severe forgetting of the split network. Liu~\etal~\cite{liu2020energy}  proposed a method for detecting out-of-distribution (OOD) data by analyzing the difference in energy scores between in-distribution (ID) and OOD data. We conducted experiments to compare the two-stage split process from CGCD with the energy-based discovery method to identify trained categories (known to the model) and untrained categories (unknown to the model).

In \Cref{fig:method_energy}, we found that the two-stage split of the CGCD struggled to identify both known and unknown categories in our proposed online learning scenario. Online learning with batch-wise data led to severe forgetting in the split network, resulting in poor detection of unknown samples. In contrast, energy-based discovery showed better performance in novel category discovery without extra parameters compared to CGCD. Moreover, as it does not require an additional learning phase, energy-based discovery enables end-to-end training. Inspired by this observation, we propose an energy-guided discovery approach for novel category discovery. To the best of our knowledge, this is the first work to utilize the energy score for novel category discovery.

During the incremental session, the network trains unlabeled batch-wise data $B^{t}=\{x_1,x_2,\dots,x_{\left|B^{t}\right|}\}$, where $t$ is the index of the batch. We classify unlabeled data $B^t$ into known data, seen data, and unseen data using both the $\theta^{\textit{off}}$ and $\theta^{on}$ through a two-stage process. 
In the first stage, we split the data into known and unknown categories. We obtain logits $z$ using the offline model $\theta^{\textit{off}}(\cdot)= g^{\textit{off}}(f^{\textit{off}}(\cdot))$ and calculate the energy scores using these logits.
The formulation for calculating the energy scores is as follows:
\begin{align}
    z^t &= \theta^{\textit{off}}(B^t) \in \mathbb{R}^{\left|B^t\right| \times \left|\mathcal{Y}_l\right|}, \\
    e^t &= \{e^t_i|e^t_i = -\mathrm{log}\sum_{j=1}^{\left|\mathcal{Y}_l\right|}\text{exp}(z_{i,j}^{t})\},
\end{align}
where $\text{exp}(\cdot)$ denotes an exponential function, $e^t$ denotes a set of energy scores, and $z_{i,j}$ denotes the logit from $i\textit{-}$th sample corresponding to the $j\textit{-}$th class.
We divide the energy scores by the Gaussian Mixture Model (GMM).
We initially set the GMM to have two clusters and use the energy score of each sample as input for clustering. We then label the cluster with a low energy score as “known” and the cluster with a high energy score as “unknown”.
This division allows us to split the unlabeled data into known and unknown data. In the second stage, we divide the unknown data into seen and unseen categories. Seen data comprise the data belonging to observed novel categories during the incremental session, while unseen data belong to undiscovered categories. To split the unknown data, we utilize the online model $\theta^{on}(\cdot)= g^{on}(f^{on}(\cdot))$ to extract energy scores. The process in the second stage is the same as the first stage, where we split the unknown data into seen and unseen categories based on their energy scores. For the initial batch of incremental sessions, the online model is identical to the offline model. Additionally, since the initial batch data is the first data of the incremental session, any data classified as unknown is assumed to be unseen data. The proposed energy-guided discovery splits unlabeled data into known, seen, and unseen categories effectively. Unlike prior methods which require sufficient data, it can identify novel categories with batch-wise data.

\subsection{Pseudo-Labeling via Variance-based Feature Augmentation}
\label{sec:vfr}
After the discovery step, we pseudo-label known, seen, and unseen data to assign labels to each sample. For known and seen data, where the model has already learned knowledge for the categories, we utilize predictions from the offline model for known data and from the online model for seen data to assign pseudo-labels, respectively. Conversely, as the unseen data belong to undiscovered categories, the online model cannot predict proper categories for them and needs to discover novel categories. Moreover, due to the given batch-wise data, discovering novel categories to assign pseudo labels is challenging.

To address this challenge, we propose variance-based feature augmentation, which temporally augments feature vectors using variational knowledge from unseen data to acquire accurate novel clusters. We obtain feature vectors of the unseen data $B_u^t$ using the online feature extractor $f^{on}(\cdot)$ denoted as $h_u^t=f^{on}(B_u^t)$. 
We calculate the variance of the unseen feature vectors and augment unseen feature vectors based on the variance. We employ each feature vector to augment features by the Gaussian distribution $\hat{h}_u^t= \{\hat{h}_u \sim \mathcal{N}(h_u,\sigma_u^2)\}, \, h_u \in h_u^t$. The formulation of the variance-based feature augmentation is as follows:
\begin{align}
    h_{\textit{aug}}^{t} = \{h_u^t;\hat{h}_u^t\},\quad \hat{h}_u^t= \{\hat{h}_u \sim \mathcal{N}(h_u,\sigma_u^2)\} \in \mathbb{R}^{(\left|B_u^t\right| \cdot K) \times d},
\end{align}
where $\hat{h}_u$ and $\sigma_u$ represent augmented feature vectors and the standard deviation of the unseen feature vectors, respectively. $K$ denotes the hyperparameter for the number of augmented feature vectors, and $h_{\textit{aug}}^{t}$ denotes the concatenation of unseen feature vectors and augmented feature vectors. Novel categories are discovered based on $h_{\textit{aug}}^{t}$. Augmented feature vectors expand the feature space with variational knowledge, which helps the model discover novel clusters.

We employ affinity propagation~\cite{frey2007clustering} to assign pseudo-labels to unseen data. Affinity propagation is a non-parametric clustering algorithm that does not require a predefined number of clusters, using the similarity matrix of the data.
Given the absence of prior knowledge of novel categories, we extend the online classifier with the estimated number of novel clusters obtained by affinity propagation. The proposed variance-based feature augmentation is a simple yet effective approach that helps the model assign effective pseudo-labels based on accurate novel category discovery.

\subsection{Online Continual Learning with Pseudo-Labeled Data}
\label{sec:inc_learn}
For online continual learning, we employ parameter-efficient tuning to learn novel representation knowledge while preserving previously learned knowledge, as it effectively mitigates catastrophic forgetting. We use cross-entropy loss with pseudo-labeled data in an online manner to capture novel knowledge. 
However, adapting the model with batch-wise data raises recency bias, where sufficient training through the given batch-wise data impedes discriminative knowledge. To address this, we propose an energy-based contrastive loss ($\mathcal{L}_{ec}$) for effective discriminative learning.
The energy-based contrastive loss focuses on inducing the online model to capture the knowledge of novel categories. It amplifies the energy from the known classifier while reducing the energy from the novel classifier. By selectively applying $\mathcal{L}_{ec}$ to novel data $B_n^t$ including seen and unseen data, it helps the online model learn novel discriminative knowledge. The formulation of $\mathcal{L}_{ec}$ is as follows: 
\begin{equation}
  \mathrm{E}(f(x);g) = -\mathrm{log} \sum_{i=1}^{\left|\mathcal{Y}\right|}\text{exp}(g_i(f(x))), \\
  \label{eq:energy}
\end{equation}
\begin{align}
    \mathcal{L}_{ec} =\frac{1}{\left|B_n^t\right|}\sum_{(x_n) \sim B_n^t} \mathrm{log}\left(1+\frac{\mathrm{E}(f^{on}(x_n);g^{new})}{\mathrm{E}(f^{on}(x_n);g^{old})}\right), \
\end{align}
where $\mathrm{E}(\cdot;\cdot)$ and $\left|\mathcal{Y}\right|$ denote energy function and the number of nodes at the classifier, respectively, $g_i(\cdot)$ represents the $i$-th node from the given classifier $g$, and $g^{old}$ and $g^{new}$ denote the nodes corresponding to known categories and discovered novel categories in the online classifier $g^{on}(\cdot)$, respectively. Eq. \ref{eq:energy} represents the calculation of the energy score given the feature vector and classifier.
The total loss $\mathcal{L}_{inc}$ for the incremental session can be summarized as follows:
\begin{align}
    \mathcal{L}_{inc}= \mathcal{L}_{ce} + \mathcal{L}_{ec},
\end{align}
where $\mathcal{L}_{ec}$ represents energy-based contrastive loss, and $\mathcal{L}_{ce}$ denotes cross-entropy loss with pseudo-labeled data.

\section{Experiments}
\subsection{Experimental Setup}
\textbf{Baselines and Implementation Details.}
Recently, many approaches have emerged in novel category discovery. However, many of them focus on self-supervised learning with non-parametric classifiers (\eg K-means)~\cite{wu2023metagcd} and rarely consider continual learning~\cite{zhao2021novel}. Some methods that consider continual learning assume prior knowledge for novel category discovery. 
Since the proposed EGD and fine-split from CGCD require a classifier to estimate the number of novel categories, non-parametric classifiers cannot be used. Additionally, existing methods not designed for continual learning are unsuitable baselines due to severe forgetting of known classes.
For a fair comparison, we mainly compared our method with FRoST~\cite{roy2022class} and CGCD~\cite{kim2023proxy}, which focus on continual learning and utilize parametric classifiers for novel category discovery.

We used ViT-B-16~\cite{dosovitskiy2020image} pre-trained on DINO-ImageNet~\cite{caron2021emerging} as the backbone network for all methods, including ours. We trained our method 30 epochs for the base session and 15 epochs for the incremental session. We utilized the AdamW optimizer with a weight decay of 1e-4 and a learning rate of 1e-3. We trained our model using a single RTX 3090 GPU with a batch size set to 64.

\noindent
\textbf{Datasets.}
We evaluated our method with state-of-the-art~(SoTA) continuous category discovery methods on three datasets: CUB200~\cite{wah2011caltech}, FGVC-Aircraft~\cite{maji2013fine}, and CIFAR-100~\cite{krizhevsky2009learning}. For continual learning, we partitioned CUB200 and FGVC-Aircraft following the same split configuration in CGCD. For CIFAR-100, we divided 80 classes for known classes and 20 classes for novel classes. 
To follow the GCD setting, we divided the training samples into labeled and unlabeled per class, maintaining a ratio of 0.8 for the labeled to 0.2 for the unlabeled.

\noindent
\textbf{Evaluation Metrics.}
We evaluated the methods by measuring clustering accuracy on test dataset $\mathcal{D}$. We utilized the hungarian algorithm~\cite{kuhn1955hungarian} to obtain the optimal permutation $h^*$ that aligns the prediction $y_i^*$ from the model with the ground truth label $y_i$:
\begin{align}
    h^* = \underset{h}{\text{arg min}} \frac{1}{|\mathcal{D}|} \sum_{i=1}^{|\mathcal{D}|} \mathbb{I}(y_i = h(y_i^*)),
\end{align}
where $|\mathcal{D}|$ is the size of the test dataset. Clustering accuracy on $\mathcal{D}$ is defined as follows:
\begin{align}
     M = \frac{1}{|\mathcal{D}|} \sum_{i=1}^{|\mathcal{D}|} \mathbb{I}(y_i = h^*({y}_i^*)). \
\end{align}
We employed this metric to measure the clustering accuracy of all classes ($M_{all}$), known classes ($M_{old}$), and unknown classes ($M_{new}$) in the cases of test dataset $\mathcal{D}^{base}\, \text{and} \,\mathcal{D}^{inc}$, respectively. Following continuous GCD approaches to evaluate the ability to maintain performance for known classes~\cite{zhang2022grow,kim2023proxy}, we employed the forgetting metric $F = M_{old}^{base} - M_{old}^{inc}$,
where $M_{old}^{base}$, $M_{old}^{inc}$ refers to known class clustering accuracy at the base session and incremental session. 
A higher clustering accuracy metric is better, and a lower forgetting metric is better.
This ensures the model is effective for real-world applications. 

Moreover, we evaluate the estimated pseudo-labels to validate their effectiveness in pseudo-labeling. Pseudo-labeling for unlabeled data is crucial as the model utilizes pseudo-labels to interpret the unlabeled data. The pseudo-labeling accuracy on the dataset $\mathcal{D}$ is defined as follows:
\begin{align}
     M = \frac{1}{|\mathcal{D}|} \sum_{i=1}^{|\mathcal{D}|} \mathbb{I}(y_i = h^*(\Tilde{{y}_i})), \
\end{align}
where $\Tilde{{y}_i}$ denotes the pseudo-label of the $i$-th sample.

\begin{table}[t]
\caption{Clustering performance comparison on CUB200, FGVC-Aircraft, and CIFAR-100. Best performance is highlighted in bold. All the baselines utilize the fine-split method from CGCD to handle the absence of prior knowledge.}
\resizebox{\columnwidth}{!}{%
\setlength{\tabcolsep}{1.5pt}
\renewcommand{\arraystretch}{1.5}
\begin{tabular}{ccccccccccccccccc}
\toprule
\multirow{2}{*}{\large Method} &  & \multicolumn{4}{c}{\scalebox{1.2}{CUB200}}      &  & \multicolumn{4}{c}{\scalebox{1.2}{FGVC-Aircraft}} &  & \multicolumn{4}{c}{\scalebox{1.2}{CIFAR-100}}     &  \\ \cline{3-6} \cline{8-11} \cline{13-16}
                        &  &  {\scalebox{1.2}{$M_{all}$}} & {\scalebox{1.2}{$M_{old}$}} & {\scalebox{1.2}{$M_{new}$}} & {\scalebox{1.2}{$F\downarrow$}}&  &  {\scalebox{1.2}{$M_{all}$}} & {\scalebox{1.2}{$M_{old}$}} & {\scalebox{1.2}{$M_{new}$}} & {\scalebox{1.2}{$F\downarrow$}} &  &  {\scalebox{1.2}{$M_{all}$}} & {\scalebox{1.2}{$M_{old}$}} & {\scalebox{1.2}{$M_{new}$}} & {\scalebox{1.2}{$F\downarrow$}} &  \\ \hline
Supervised              &  & 72.51 & 70.68 & 79.66   & 11.91  &  & 66.25  & 63.62 & 76.76   & 11.03  &  & 81.34 & 79.48 & 88.80   & 10.07  &  \\ \hdashline
Fine-Tuning             &  & 1.96 & 1.94 & 2.01    & 80.52  &  & 2.06  & 1.30 & 5.10    & 73.35   &  & 3.68 & 3.24 & 5.45    & 86.26  &  \\
FRoST {[}ECCV’ 22{]}    &  & 23.98 & 28.72 & 5.45    & 50.85  &  & 14.44  & 15.65 & 9.60    & 48.31  &  & 6.13  & 6.21  & 5.80    & 69.24  &  \\
CGCD {[}ICCV’ 23{]}      &  & 40.40 & 47.88 & 11.18   & 35.61  &  & 14.08  & 15.67 & 7.75    & 60.00  &  & 2.64  & 2.76  & 2.15    & 86.74  &  \\ \hline
DEAN w/ FS              &  & 55.57 & 65.37 & 17.29 & 17.62    &  &   40.32     &  49.14     &   5.10      &   25.43     &  &    52.27   &    62.11   &  12.90       &    27.55    &  \\
\textbf{DEAN (Ours)}             &  & \textbf{59.56} & \textbf{66.33} & \textbf{33.11}   & \textbf{16.26}  &  & \textbf{46.44}  & \textbf{53.56} & \textbf{17.99}   & \textbf{20.97}  &  & \textbf{62.34} & \textbf{70.34} & \textbf{30.35}   & \textbf{19.17}  &\\
\bottomrule
\end{tabular}%
}
\label{tab:main}
\end{table}

\begin{table}[t]
\begin{minipage}{0.5\columnwidth} \centering
\captionof{table}{Ablation experiment on CUB200. The baseline denotes fine-tuning with the fine split from the CGCD.}
\label{tab:ablation}
\resizebox{\columnwidth}{!}{%
\setlength{\tabcolsep}{2pt}
\renewcommand{\arraystretch}{1.35}
\begin{tabular}{lcccccl}
\toprule
\multicolumn{1}{c}{\multirow{2}{*}{Ablation}}            &  & \multicolumn{5}{c}{CUB200}                                                                                        \\ \cline{2-6}
                                   &  & \multicolumn{1}{c}{{$M_{all}$}} & \multicolumn{1}{c}{$M_{old}$} & \multicolumn{1}{c}{$M_{new}$} & \multicolumn{1}{c}{$F\downarrow$} &  \\ \hline
Baseline                           &  & 1.96                    & 1.94                      & 2.01                        & 80.52                      &  \\ \hline
\textbf{+} LoRA                             &  & 49.33                   & 60.35                     & 6.24                        & 22.24                      &  \\
\textbf{+} EGD          &  & 54.97                   & 62.98                     & 23.67                       & 19.61                      &  \\
\textbf{+} EC Loss    &  & 57.37                   & 64.98                     & 27.63                       & 17.61                      &  \\ \hline
\textbf{+ VFA (Ours)} &  & \textbf{59.56}                   & \textbf{66.33}                     & \textbf{33.11}                       & \textbf{16.26}                      & \\
\bottomrule
\end{tabular}%
}
\end{minipage}
\hfill
\begin{minipage}{0.45\columnwidth}
\centering
\captionof{table}{Ablation study for the hyperparameter $\mathrm{K}$ of variance-based feature augmentation on CUB200.}
\label{tab:ablation_vfa}
\resizebox{\columnwidth}{!}{%
\setlength{\tabcolsep}{8.pt}
\renewcommand{\arraystretch}{1.2}
\begin{tabular}{cccc}
\toprule
\multirow{2}{*}{Ablation} & \multicolumn{3}{c}{CUB200} \\ \cline{2-4}
                          & $M_{all}^{PS}$ & $M_{old}^{PS}$ & $M_{new}^{PS}$ \\ \cline{1-4}
K=0                       & 57.31    & 85.19      & 46.47        \\ \cline{1-4}
K=1                       & 64.27    & 88.96      & 53.74        \\
K=3                       & 75.87    & 93.43      & 64.48        \\
\textbf{K=5}              &\textbf{76.93}    & \textbf{94.05}      & \textbf{65.75}        \\
K=7                       & 75.44    & 93.41      & 64.07        \\
K=9                       & 75.22    & 93.5       & 63.36   \\
\bottomrule
\end{tabular}%
}
\end{minipage}
\end{table}

\subsection{Comparison with Baselines}
\label{sec:baselines}
We compared the clustering performance of our proposed method, DEAN, with SoTA approaches, as shown in \Cref{tab:main}. Since we considered an online scenario for category discovery, traditionally approached using offline methods, there is no previous work for comparison. Therefore, for a fair comparison, our study adapts offline continuous NCD and GCD methods into OCGCD scenarios. The experiment includes two SoTA continuous category discovery methods: FRoST~\cite{roy2022class} and CGCD~\cite{kim2023proxy}. In addition, we include a supervised learning method as an upper bound and a fine-tuning method as a performance baseline. 
Note that in our proposed scenario, OCGCD, all baselines are not provided with prior knowledge about novel categories.
Instead, we employed fine-split method from CGCD~\cite{kim2023proxy} which consists of two stages: initial split and fine split. During the initial split stage, unlabeled data is roughly divided into known and unknown categories based on cosine similarity. In the fine split stage, a split model is trained to distinguish between known and unknown data.
The fine-split method estimates the number of novel categories conveying information about novel categories to the comparison methods.

The results shows that our method, DEAN, recorded a robust clustering performance, significantly surpassing existing continuous NCD and GCD methods. DEAN recorded consistently high metrics across $M_{all},\, M_{old}$, and $M_{new}$ while keeping forgetting metric $F$ substantially low. 
This denotes our method can learn novel knowledge without forgetting.
Moreover, it is noteworthy that DEAN w/~FS which denotes the adoption of fine-split from CGCD into DEAN instead of energy-guided discovery~(EGD) also showed notable performance compared to the baselines. But, DEAN achieved significant performance gains across all datasets as the proposed EGD is applied.
In contrast, existing continual methods showed shortcomings in clustering performance. Specifically, we observed extremely low clustering performance on CIFAR-100 dataset, low $ M_{all}$, $M_{old}$, and $M_{new}$ values, coupled with high $F$. On CIFAR-100 dataset, DEAN surpassed the most recent method CGCD assuming no prior knowledge for novel categories by 59.7\%, 67.58\%, 28.2\%, and 67.57\% for $ M_{all}$, $M_{old}, M_{new}$, and $F$.

Such observations shows that while current GCD methods are vulnerable to online settings, DEAN can manage the balance required in the OCGCD scheme of retaining old knowledge while integrating the new. We believe this results shows the suitability of DEAN for real-world applications, where data is never static. We  empirically confirmed that DEAN can evolve with the data it encounters, overcoming the challenges presented in the OCGCD scheme. 

\begin{table}[t]
\begin{minipage}{0.48\columnwidth} \centering
\captionof{table}{Ablation study for the cluster accuracy of pseudo-labels.}
\label{tab:ablation_ps}
\resizebox{\columnwidth}{!}{%
\setlength{\tabcolsep}{4.pt}
\renewcommand{\arraystretch}{1.3}
\begin{tabular}{cccc}
\toprule
\multirow{2}{*}{Method} & \multicolumn{3}{c}{CUB200} \\ \cline{2-4}
 & {$M_{all}^{PS}$} & {$M_{old}^{PS}$} & {$M_{new}^{PS}$} \\ \hline
\multicolumn{1}{l}{CGCD [ICCV’ 23]} &  61.88  & 81.07    & 58.68      \\ \hline
\multicolumn{1}{l}{DEAN w/o ($\mathcal{L}_{ec}$, VFA)} &  57.23  & 85.11    & 45.80      \\
\multicolumn{1}{l}{DEAN w/o VFA}    &  57.31  & 85.19    & 46.47      \\ \hline
\multicolumn{1}{l}{\textbf{DEAN (Ours)}} &  \textbf{76.93}    & \textbf{94.05}      & \textbf{65.75}     \\
\bottomrule
\end{tabular}%
\label{tab:ps_acc}
}
\end{minipage}
\hfill
\begin{minipage}{0.48\columnwidth} 
\centering
\captionof{table}{Ablation experiment of parameter-efficient tuning on CUB200.}
\label{tab:ablation_pet}
\resizebox{\columnwidth}{!}{%
\setlength{\tabcolsep}{1.mm}
\renewcommand{\arraystretch}{1.6}
\begin{tabular}{cclcccc}
\toprule
\multicolumn{3}{c}{\multirow{2}{*}{Method}} & \multicolumn{4}{c}{CUB200} \\ \cline{4-7} 
\multicolumn{3}{c}{} & $M_{all}$ & $M_{old}$ & $M_{new}$ & $F\downarrow$ \\ \hline
\multirow{3}{*}{\begin{tabular}[c]{@{}c@{}}DEAN\\ (Ours)\end{tabular}} &  & Adapter~\cite{chen2022adaptformer} & 56.96 & 64.75 & 26.53 & 17.64 \\
 &  & Prefix~\cite{li2021prefix} & 58.54 & 66.29 & 28.28 & 16.30 \\ \cline{3-7} 
 &  & \textbf{LoRA}~\cite{hu2021lora} & \textbf{59.56} & \textbf{66.33} & \textbf{33.11} & \textbf{16.26} \\ \bottomrule
\end{tabular}%
}
\end{minipage}

\end{table}

\subsection{Ablation Study}
The experimental results in \Cref{sec:baselines} show that our DEAN achieved significant performance in OCGCD scenarios. Through further experiments on the CUB200, we provide a detailed analysis of the components proposed in our method.

\noindent
\textbf{Effectiveness of Components.}
We validate the effectiveness of adding each component of our method to the baseline model. The baseline denotes fine-tuning the backbone model with a data-split following settings in CGCD. As shown in \Cref{tab:ablation}, results reveal that all components have an important impact on performance. Employing LoRA to baseline shows significant improvement in $M_{old},\, F$. This shows that our application of LoRA on the frozen model at incremental session effectively mitigates forgetting. While LoRA shows only a slight increase in $M_{new}$, additionally employing energy-guided discovery (EGD) and EC Loss markedly improves $M_{new}$. This confirms that adopting energy guidance will enhance the ability of the model to differentiate between old and new categories. Finally, the integration of VFA further boosts the model performance on $M_{new}$. This improvement reveals the importance of accurate pseudo-labeling since VFA is a technique applied after training. Each component is crucial, incrementally building upon the last to solidify the model's performance in our online category discovery scenario. 

\noindent
\textbf{Hyperparameter Analysis of VFA.} 
Variance-based Feature Augmentation (VFA) is our proposed method of temporarily augmenting feature vectors to acquire accurate pseudo-labeling. Our analysis evaluated the effect of hyperparameter $\mathrm{K}$, the number of augmented feature vectors, on the accuracy of pseudo-labels. From \Cref{tab:ablation_vfa} we observed that employment of VFA is beneficial, increasing novel category pseudo-label accuracy from 46.47\% to 65.75\%. Among VFA employment, we observed that $\mathrm{K}=5$ yields the best performance across all metrics. When $\mathrm{K}$ is less than 5, the smaller $\mathrm{K}$ worsened performance. Conversely, values of $\mathrm{K}$ greater than 5 did not benefit the model. We analyzed that since VFA augments features based on variance, as the number of $\mathrm{K}$ increases, unseen samples can be clustered together without being distinguished.

\noindent
\textbf{Pseudo-Labeling Accuracy.}
Accurate pseudo-labeling for clustering is important because it directly influences the performance of the model in incremental learning. Shown in \Cref{tab:ps_acc}, our proposed method, DEAN, significantly enhanced the clustering accuracy of pseudo-labels, achieving a 13.12\% improvement in $M_{all}^{PS}$ over the most recent method CGCD. Through the ablation study, we investigated the contributions of each step in DEAN. When DEAN adopts only parameter-efficient tuning, excluding EC loss and VFA, we observed a notable declination in $M_{new}^{PS}$. While each ablation step showed gradually accuracy enhancement, the application of VFA elevates the performance of DEAN above CGCD. Thus, the experimental result revealed that VFA played a vital role in distinguishing novel categories and confirming the effectiveness of our method.

\noindent
\textbf{Parameter Efficient Tuning.}
The ablation study in \Cref{tab:ablation_pet} assesses the impact of parameter-efficient tuning (PET) methods on the CUB200 dataset. Our method can employ various tuning methods such as Adapter~\cite{chen2022adaptformer} or Prefix-tuning~\cite{li2021prefix}, and in this experiment, we compared modules with LoRA~\cite{hu2021lora}, Adapter and Prefix-tuning. We set the size of all PET modules size to 5 and integrated them into the last 5 layers of ViT. Overall improvements are shown by adopting PET over existing methods reported in \Cref{tab:main}. Among PET methods, LoRA showed the highest clustering accuracy. The performance enhancement of LoRA may stem from its fast adaptation, which is highly effective in online continual learning scenarios that often depend on fewer samples for training.

\label{sec:blind}

\section{Conclusion}
In this paper, we introduced a novel scenario for category discovery, Online Continuous Generalized Category Discovery, which accounts for the dynamic nature of real-world data streams, leading to severe forgetting and poor novel category discovery. In addition, we proposed a novel framework, DEAN, which utilizes energy-based novel category discovery for the first time in this field and introduces variance-based feature augmentation to enhance accurate pseudo-labeling. Our DEAN demonstrated promising performance in the proposed novel scenario. However, DEAN could not guarantee inference performance midway through training. In future work, we plan to explore an online continual framework that can ensure inference performance during training through fast adaptation.

\section*{Acknowledgements}
This work was supported by MSIT (Ministry of Science and ICT), Korea, under the ITRC (Information Technology Research Center) support program (IITP-2024-RS-2023-00258649) supervised by the IITP (Institute for Information \& Communications Technology Planning \& Evaluation), and in part by the IITP grant funded by the Korea Government (MSIT) (Artificial Intelligence Innovation Hub) under Grant 2021-0-02068, and by the IITP grant funded by the Korea government (MSIT) (No.RS-2022-00155911, Artificial Intelligence Convergence Innovation Human Resources Development (Kyung Hee University)).


%
%
\nocite{seo2024generative,park2024pre,choe2024open}

\bibliographystyle{splncs04}
\bibliography{main}

\begin{thebibliography}{10}
\providecommand{\url}[1]{\texttt{#1}}
\providecommand{\urlprefix}{URL }
\providecommand{\doi}[1]{https://doi.org/#1}

\bibitem{aljundi2019online}
Aljundi, R., Belilovsky, E., Tuytelaars, T., Charlin, L., Caccia, M., Lin, M., Page-Caccia, L.: Online continual learning with maximal interfered retrieval. Advances in Neural Information Processing Systems  \textbf{32} (2019)

\bibitem{arpit2017closer}
Arpit, D., Jastrz{\k{e}}bski, S., Ballas, N., Krueger, D., Bengio, E., Kanwal, M.S., Maharaj, T., Fischer, A., Courville, A., Bengio, Y., et~al.: A closer look at memorization in deep networks. In: International conference on machine learning. pp. 233--242. PMLR (2017)

\bibitem{caron2021emerging}
Caron, M., Touvron, H., Misra, I., J\'egou, H., Mairal, J., Bojanowski, P., Joulin, A.: Emerging properties in self-supervised vision transformers. In: Proceedings of the International Conference on Computer Vision (ICCV) (2021)

\bibitem{chen2022adaptformer}
Chen, S., Ge, C., Tong, Z., Wang, J., Song, Y., Wang, J., Luo, P.: Adaptformer: Adapting vision transformers for scalable visual recognition. Advances in Neural Information Processing Systems  \textbf{35},  16664--16678 (2022)

\bibitem{choe2024open}
Choe, S.A., Shin, A.H., Park, K.H., Choi, J., Park, G.M.: Open-set domain adaptation for semantic segmentation. In: Proceedings of the IEEE/CVF Conference on Computer Vision and Pattern Recognition. pp. 23943--23953 (2024)

\bibitem{dosovitskiy2020image}
Dosovitskiy, A., Beyer, L., Kolesnikov, A., Weissenborn, D., Zhai, X., Unterthiner, T., Dehghani, M., Minderer, M., Heigold, G., Gelly, S., et~al.: An image is worth 16x16 words: Transformers for image recognition at scale. In: International Conference on Learning Representations (2020)

\bibitem{fini2021unified}
Fini, E., Sangineto, E., Lathuiliere, S., Zhong, Z., Nabi, M., Ricci, E.: A unified objective for novel class discovery. In: Proceedings of the IEEE/CVF International Conference on Computer Vision. pp. 9284--9292 (2021)

\bibitem{french1999catastrophic}
French, R.M.: Catastrophic forgetting in connectionist networks. Trends in cognitive sciences  \textbf{3}(4),  128--135 (1999)

\bibitem{frey2007clustering}
Frey, B.J., Dueck, D.: Clustering by passing messages between data points. science  \textbf{315}(5814),  972--976 (2007)

\bibitem{guo2022online}
Guo, Y., Liu, B., Zhao, D.: Online continual learning through mutual information maximization. In: International Conference on Machine Learning. pp. 8109--8126. PMLR (2022)

\bibitem{gutmann2010noise}
Gutmann, M., Hyv{\"a}rinen, A.: Noise-contrastive estimation: A new estimation principle for unnormalized statistical models. In: Proceedings of the thirteenth international conference on artificial intelligence and statistics. pp. 297--304. JMLR Workshop and Conference Proceedings (2010)

\bibitem{han2020automatically}
Han, K., Rebuffi, S.A., Ehrhardt, S., Vedaldi, A., Zisserman, A.: Automatically discovering and learning new visual categories with ranking statistics. arXiv preprint arXiv:2002.05714  (2020)

\bibitem{han2019learning}
Han, K., Vedaldi, A., Zisserman, A.: Learning to discover novel visual categories via deep transfer clustering. In: Proceedings of the IEEE/CVF International Conference on Computer Vision. pp. 8401--8409 (2019)

\bibitem{he2020incremental}
He, J., Mao, R., Shao, Z., Zhu, F.: Incremental learning in online scenario. In: Proceedings of the IEEE/CVF conference on computer vision and pattern recognition. pp. 13926--13935 (2020)

\bibitem{hsu2017learning}
Hsu, Y.C., Lv, Z., Kira, Z.: Learning to cluster in order to transfer across domains and tasks. arXiv preprint arXiv:1711.10125  (2017)

\bibitem{hsu2019multi}
Hsu, Y.C., Lv, Z., Schlosser, J., Odom, P., Kira, Z.: Multi-class classification without multi-class labels. arXiv preprint arXiv:1901.00544  (2019)

\bibitem{hu2021lora}
Hu, E.J., Wallis, P., Allen-Zhu, Z., Li, Y., Wang, S., Wang, L., Chen, W., et~al.: Lora: Low-rank adaptation of large language models. In: International Conference on Learning Representations (2021)

\bibitem{jia2021joint}
Jia, X., Han, K., Zhu, Y., Green, B.: Joint representation learning and novel category discovery on single-and multi-modal data. In: Proceedings of the IEEE/CVF International Conference on Computer Vision. pp. 610--619 (2021)

\bibitem{joseph2022novel}
Joseph, K., Paul, S., Aggarwal, G., Biswas, S., Rai, P., Han, K., Balasubramanian, V.N.: Novel class discovery without forgetting. In: European Conference on Computer Vision. pp. 570--586. Springer (2022)

\bibitem{khosla2020supervised}
Khosla, P., Teterwak, P., Wang, C., Sarna, A., Tian, Y., Isola, P., Maschinot, A., Liu, C., Krishnan, D.: Supervised contrastive learning. Advances in neural information processing systems  \textbf{33},  18661--18673 (2020)

\bibitem{kim2022splitnet}
Kim, D., Ryoo, K., Cho, H., Kim, S.: Splitnet: learnable clean-noisy label splitting for learning with noisy labels. arXiv preprint arXiv:2211.11753  (2022)

\bibitem{kim2023proxy}
Kim, H., Suh, S., Kim, D., Jeong, D., Cho, H., Kim, J.: Proxy anchor-based unsupervised learning for continuous generalized category discovery. In: Proceedings of the IEEE/CVF International Conference on Computer Vision. pp. 16688--16697 (2023)

\bibitem{koh2021online}
Koh, H., Kim, D., Ha, J.W., Choi, J.: Online continual learning on class incremental blurry task configuration with anytime inference. arXiv preprint arXiv:2110.10031  (2021)

\bibitem{krizhevsky2009learning}
Krizhevsky, A., Hinton, G., et~al.: Learning multiple layers of features from tiny images  (2009)

\bibitem{kuhn1955hungarian}
Kuhn, H.W.: The hungarian method for the assignment problem. Naval research logistics quarterly  \textbf{2}(1-2),  83--97 (1955)

\bibitem{li2020dividemix}
Li, J., Socher, R., Hoi, S.C.: Dividemix: Learning with noisy labels as semi-supervised learning. arXiv preprint arXiv:2002.07394  (2020)

\bibitem{li2021prefix}
Li, X.L., Liang, P.: Prefix-tuning: Optimizing continuous prompts for generation. In: Proceedings of the 59th Annual Meeting of the Association for Computational Linguistics and the 11th International Joint Conference on Natural Language Processing (Volume 1: Long Papers). pp. 4582--4597 (2021)

\bibitem{liu2020energy}
Liu, W., Wang, X., Owens, J., Li, Y.: Energy-based out-of-distribution detection. Advances in neural information processing systems  \textbf{33},  21464--21475 (2020)

\bibitem{liu2022residual}
Liu, Y., Tuytelaars, T.: Residual tuning: Toward novel category discovery without labels. IEEE Transactions on Neural Networks and Learning Systems  (2022)

\bibitem{mai2022online}
Mai, Z., Li, R., Jeong, J., Quispe, D., Kim, H., Sanner, S.: Online continual learning in image classification: An empirical survey. Neurocomputing  \textbf{469},  28--51 (2022)

\bibitem{maji2013fine}
Maji, S., Rahtu, E., Kannala, J., Blaschko, M., Vedaldi, A.: Fine-grained visual classification of aircraft. arXiv preprint arXiv:1306.5151  (2013)

\bibitem{mcclelland1995there}
McClelland, J.L., McNaughton, B.L., O'Reilly, R.C.: Why there are complementary learning systems in the hippocampus and neocortex: insights from the successes and failures of connectionist models of learning and memory. Psychological review  \textbf{102}(3), ~419 (1995)

\bibitem{mccloskey1989catastrophic}
McCloskey, M., Cohen, N.J.: Catastrophic interference in connectionist networks: The sequential learning problem. In: Psychology of learning and motivation, vol.~24, pp. 109--165. Elsevier (1989)

\bibitem{moon2023online}
Moon, J.Y., Park, K.H., Kim, J.U., Park, G.M.: Online class incremental learning on stochastic blurry task boundary via mask and visual prompt tuning. In: Proceedings of the IEEE/CVF International Conference on Computer Vision. pp. 11731--11741 (2023)

\bibitem{nishi2021augmentation}
Nishi, K., Ding, Y., Rich, A., Hollerer, T.: Augmentation strategies for learning with noisy labels. In: Proceedings of the IEEE/CVF Conference on Computer Vision and Pattern Recognition. pp. 8022--8031 (2021)

\bibitem{park2024pre}
Park, K.H., Song, K., Park, G.M.: Pre-trained vision and language transformers are few-shot incremental learners. In: Proceedings of the IEEE/CVF Conference on Computer Vision and Pattern Recognition. pp. 23881--23890 (2024)

\bibitem{roy2022class}
Roy, S., Liu, M., Zhong, Z., Sebe, N., Ricci, E.: Class-incremental novel class discovery. In: European Conference on Computer Vision. pp. 317--333. Springer (2022)

\bibitem{seo2024generative}
Seo, J., Lee, S.H., Lee, T.Y., Moon, S., Park, G.M.: Generative unlearning for any identity. In: Proceedings of the IEEE/CVF Conference on Computer Vision and Pattern Recognition. pp. 9151--9161 (2024)

\bibitem{troisemaine2023novel}
Troisemaine, C., Lemaire, V., Gosselin, S., Reiffers-Masson, A., Flocon-Cholet, J., Vaton, S.: Novel class discovery: an introduction and key concepts. arXiv preprint arXiv:2302.12028  (2023)

\bibitem{vaze2022generalized}
Vaze, S., Han, K., Vedaldi, A., Zisserman, A.: Generalized category discovery. In: Proceedings of the IEEE/CVF Conference on Computer Vision and Pattern Recognition. pp. 7492--7501 (2022)

\bibitem{wah2011caltech}
Wah, C., Branson, S., Welinder, P., Perona, P., Belongie, S.: The caltech-ucsd birds-200-2011 dataset  (2011)

\bibitem{wu2023metagcd}
Wu, Y., Chi, Z., Wang, Y., Feng, S.: Metagcd: Learning to continually learn in generalized category discovery. In: Proceedings of the IEEE/CVF International Conference on Computer Vision. pp. 1655--1665 (2023)

\bibitem{zhang2021understanding}
Zhang, C., Bengio, S., Hardt, M., Recht, B., Vinyals, O.: Understanding deep learning (still) requires rethinking generalization. Communications of the ACM  \textbf{64}(3),  107--115 (2021)

\bibitem{zhang2022grow}
Zhang, X., Jiang, J., Feng, Y., Wu, Z.F., Zhao, X., Wan, H., Tang, M., Jin, R., Gao, Y.: Grow and merge: A unified framework for continuous categories discovery. Advances in Neural Information Processing Systems  \textbf{35},  27455--27468 (2022)

\bibitem{zhao2021novel}
Zhao, B., Han, K.: Novel visual category discovery with dual ranking statistics and mutual knowledge distillation. Advances in Neural Information Processing Systems  \textbf{34},  22982--22994 (2021)

\bibitem{zhong2021neighborhood}
Zhong, Z., Fini, E., Roy, S., Luo, Z., Ricci, E., Sebe, N.: Neighborhood contrastive learning for novel class discovery. In: Proceedings of the IEEE/CVF conference on computer vision and pattern recognition. pp. 10867--10875 (2021)

\bibitem{zhong2021openmix}
Zhong, Z., Zhu, L., Luo, Z., Li, S., Yang, Y., Sebe, N.: Openmix: Reviving known knowledge for discovering novel visual categories in an open world. In: Proceedings of the IEEE/CVF Conference on Computer Vision and Pattern Recognition. pp. 9462--9470 (2021)

\end{thebibliography}
\end{document}


\makeatletter
\renewcommand*{\@fnsymbol}[1]{\ensuremath{\ifcase#1\or \dagger\or \ddagger\or
   \mathsection\or \mathparagraph\or \|\or **\or \dagger\dagger
   \or \ddagger\ddagger \else\@ctrerr\fi}}
\makeatother
\newcommand*\samethanks[1][\value{footnote}]{\footnotemark[#1]}
\title{Online Continuous Generalized Category Discovery \\ {\large - Supplementary Materials -}} 
\titlerunning{Online Continuous Generalized Category Discovery}

\author{Keon-Hee Park\inst{1}\orcidlink{0009-0008-3654-812X} \and
Hakyung Lee\inst{2} \orcidlink{0009-0001-7600-9760} \and
Kyungwoo Song\inst{2}\thanks{Corresponding authors} \orcidlink{0000-0003-0082-4280} \and
Gyeong-Moon~Park\inst{1}\samethanks[1]\orcidlink{0000-0003-4011-9981}}

\authorrunning{K.-H. Park et al.}

\institute{Kyung Hee University, Republic of Korea \\
\email{\{khpark,gmpark\}@khu.ac.kr} \and
Yonsei University, Republic of Korea \\
\email{\{hakyunglee0417, kyungwoo.song\}@yonsei.ac.kr}}

\maketitle
\appendix

\begin{table}[h]
\centering
\begin{minipage}{\columnwidth} 
\captionof{table}{Dataset configuration for the proposed online continuous 
\label{tab:sup_dataset}generalized category discovery. S.Ratio denotes the ratio of samples.}
\resizebox{\columnwidth}{!}{%
\setlength{\tabcolsep}{1.pt}
\renewcommand{\arraystretch}{1.2}
\begin{tabular}{ccccc}
\toprule
\multirow{2}{*}{Dataset} & \multirow{2}{*}{All Classes} & & Base Session         & Inc Session (Known + Novel) \\ \cline{4-5}
                         &                             & & \textbf{\#} of Classes (S.Ratio) &\textbf{\#} of Classes (S.Ratio)        \\ \hline
CUB200                   & 200 &                          & 160 (0.8)            & 160 (0.2) + 40 (1.0)        \\
FGVC-Aircraft            & 100 &                          & 80 (0.8)             & 80 (0.2) + 20 (1.0)         \\
CIFAR-100                & 100 &                          & 80 (0.8)             & 80 (0.2) + 20 (1.0)        \\
\bottomrule
\end{tabular}%
}
\end{minipage}
\end{table}

\section{Further Implementation Details}
To clearly define the scenario configuration, \Cref{tab:sup_dataset} demonstrates the data configuration for the proposed online continuous generalized category discovery. The incremental session comprised samples from both known categories and novel categories. We conducted 3 simulations for all experiments using different random seeds and reported the averages.

\section{Details of Comparison Methods}

\subsubsection{FRoST} is designed to address class incremental novel category discovery, where the labeled dataset and unlabeled dataset have no overlap. FRoST utilizes an auxiliary classifier to pseudo-label novel samples and requires prior knowledge of the novel categories to expand the classifier and initialize the auxiliary classifier. In our experiments, we refrained from providing prior knowledge to ensure a fair comparison. Instead, we relied on calculating the similarity between features and classifier nodes to discover novel categories using affinity propagation.

\subsubsection{CGCD} CGCD proposed continuous generalized category discovery, where the model continuously fine-tunes unlabeled datasets after training on a labeled dataset. CGCD assumes the overlapping between labeled and unlabeled datasets and does not require prior knowledge of novel categories. And, CGCD employs the proxy anchor-based contrastive loss for effective deep metric learning and requires a split network to categorize unlabeled data into old and novel categories.

\begin{table}[t]
\centering
\begin{minipage}{0.8\columnwidth} 
\caption{Further study of the layers containing LoRA on CUB200.}
\resizebox{\columnwidth}{!}{%
\setlength{\tabcolsep}{12.pt}
\renewcommand{\arraystretch}{1.}
\begin{tabular}{ccccc}
\toprule
\multirow{2}{*}{\begin{tabular}[c]{@{}c@{}}Layers\\ From $\rightarrow$ To\end{tabular}} & \multicolumn{4}{c}{CUB200}                                       \\ \cline{2-5}
                        & $M_{all}$          & $M_{old}$        & $M_{new}$      & $F\downarrow$            \\ \hline
1 $\rightarrow$ 5                  & 55.90 & 65.18 & 22.97 & 17.61          \\
3 $\rightarrow$ 7                  & \textbf{56.01} & \textbf{65.33} & 21.36 & \textbf{17.05}          \\
5 $\rightarrow$ 9                  & 54.58 & 63.66 & 22.71 & 18.79          \\
7 $\rightarrow$ 11                 & 55.19 & 64.19 & 22.86 & 18.34          \\ \hline
\textbf{8 $\rightarrow$ 12}        & 54.97 & 62.98 & \textbf{23.67} & 19.61 \\
\bottomrule
\end{tabular}%
\label{tab:sup_pet}
}
\end{minipage}
\end{table}
\newcolumntype{g}{>{\columncolor{Gray}}c}
\begin{table}[t]
\centering
\definecolor{lightgray}{gray}{0.9}
\begin{minipage}{0.9\columnwidth}
\captionof{table}{Further experiment to evaluate the proposed online continuous generalized category discovery in various settings.}
\label{tab:ocgcd_various}
\resizebox{\columnwidth}{!}{%
\setlength{\tabcolsep}{5.pt}
\renewcommand{\arraystretch}{1.2}
\begin{tabular}{ccccccc}
\toprule 
\multicolumn{1}{c}{\multirow{2}{*}{\begin{tabular}[c]{@{}c@{}}{\large \text{[}Base / Incremental Sessions]} \\{\normalsize \text{[}\textbf{\#} of Classes(S.Ratio)]} \end{tabular}}} & \multicolumn{1}{c}{\multirow{2}{*}{{\large Method}}} & \multicolumn{1}{c}{} & \multicolumn{4}{c}{{\large CUB200}} \\ \cline{4-7}
 \multicolumn{1}{c}{} & \multicolumn{1}{c}{} & \multicolumn{1}{c}{} & \multicolumn{1}{c}{{\large $M_{all}$}} & \multicolumn{1}{c}{{\large $M_{old}$}} & \multicolumn{1}{c}{{\large $M_{new}$}} & \multicolumn{1}{c}{{\large $F\downarrow$}} \\ \hline 
\rowcolor{lightgray} & FRoST [ECCV' 22] & & 43.61 & 47.32 & 10.71 & 33.63 \\ 
\rowcolor{lightgray} & CGCD [ICCV' 23] & & 61.22 & 66.75 & 12.24 & 18.33 \\ 
\rowcolor{lightgray}\multirow{-3}{*}{\normalsize \text{[}180(0.9) / 180(0.1) + 20(1.0)]} & \textbf{DEAN (Ours)} & & \textbf{65.60} & \textbf{68.92} & \textbf{36.22} & \textbf{14.46} \\
\multirow{3}{*}{\normalsize \text{[}160(0.8) / 160(0.2) + 40(1.0)]} & FRoST [ECCV' 22] &  & 23.98 & 28.72 & 5.45 & 50.85 \\
 & CGCD [ICCV' 23] &  & 40.40 & 47.88 & 11.18 & 35.61 \\
 & \textbf{DEAN (Ours)} &  & \textbf{59.56} & \textbf{66.33} & \textbf{33.11} & \textbf{16.26} \\
 \rowcolor{lightgray} & FRoST [ECCV' 22] & & 32.61 & 44.64 & 5.16 & 34.89 \\ 
\rowcolor{lightgray} & CGCD [ICCV' 23] &  & 32.78 & 45.77 & 3.16 & 37.00 \\
\rowcolor{lightgray}\multirow{-3}{*}{\normalsize \text{[}140(0.7) / 140(0.3) + 60(1.0)]} & \textbf{DEAN (Ours)} &  & \textbf{54.60} & \textbf{64.62} & \textbf{26.36} & \textbf{17.21} \\
\multirow{3}{*}{\normalsize \text{[}120(0.6) / 120(0.4) + 80(1.0)]} & FRoST [ECCV' 22] &  & 31.00 & 47.74 & 6.71 & 30.10 \\ 
 & CGCD [ICCV' 23] &  & 27.18 & 41.94 & 5.77 & 40.57 \\ 
 & \textbf{DEAN (Ours)} &  & \textbf{47.76} & \textbf{64.00} & \textbf{23.77} & \textbf{17.22} \\
\rowcolor{lightgray} & FRoST [ECCV' 22] &  & 27.12 & 50.23 & 4.53 & 28.84 \\
\rowcolor{lightgray} & CGCD [ICCV' 23] &  & 21.03 & 38.90 & 3.56 & 43.85 \\
 \rowcolor{lightgray}\multirow{-3}{*}{\normalsize \text{[}100(0.5) / 100(0.5) + 100(1.0)]} & \textbf{DEAN (Ours)} &  & \textbf{43.46} & \textbf{63.99} & \textbf{19.21} & \textbf{17.26} \\
\multirow{3}{*}{\normalsize \text{[}60(0.3) / 60(0.7) + 140(1.0)]} & FRoST [ECCV' 22] &  & 18.78 & 53.86 & 4.39 & 20.05 \\
 & CGCD [ICCV' 23] &  & 13.73 & 31.76 & 6.33 & 45.82 \\
 & \textbf{DEAN (Ours)} &  & \textbf{27.27} & \textbf{56.88} & \textbf{15.12} & \textbf{19.38} \\
 \bottomrule
\end{tabular}%
}
\end{minipage}
\end{table}

\section{Further Analysis for the PET}
We conducted an ablation study to determine the optimal position for parameter-efficient tuning. As shown in~\Cref{tab:sup_pet}, while training LoRA from the 3rd layer to the 7th layer ($3\rightarrow7$) effectively mitigated severe forgetting, adopting these layers resulted in poor performance in novel category clustering. However, inserting LoRA into the last five layers ($8 \rightarrow 12$) recorded the highest performance in novel category clustering. Based on this observation, we selected the last five layers to facilitate learning novel categories.

\section{Qualitative Evaluation of the Proposed Online Continuous Generalized Category Discovery}
To evaluate the robustness of the proposed method, DEAN, we conducted experiments under various OCGCD settings. We scaled down the number of classes and samples in the base session following a specific ratio. As shown in \Cref{tab:ocgcd_various}, decreasing the number of classes and samples in the base session made the model struggle to discover novel categories. FRoST and CGCD suffered severe performance declination due to the decreasing proportions of classes and samples in the base session. However, DEAN consistently achieved outstanding performance across all measurements. This experimental result demonstrates that our method can maintain performance even when the model cannot be trained with sufficient data and classes during the base session.
\begin{figure}[t]
\begin{subfigure}{0.32\columnwidth}
    \centering
    \includegraphics[width=\columnwidth]{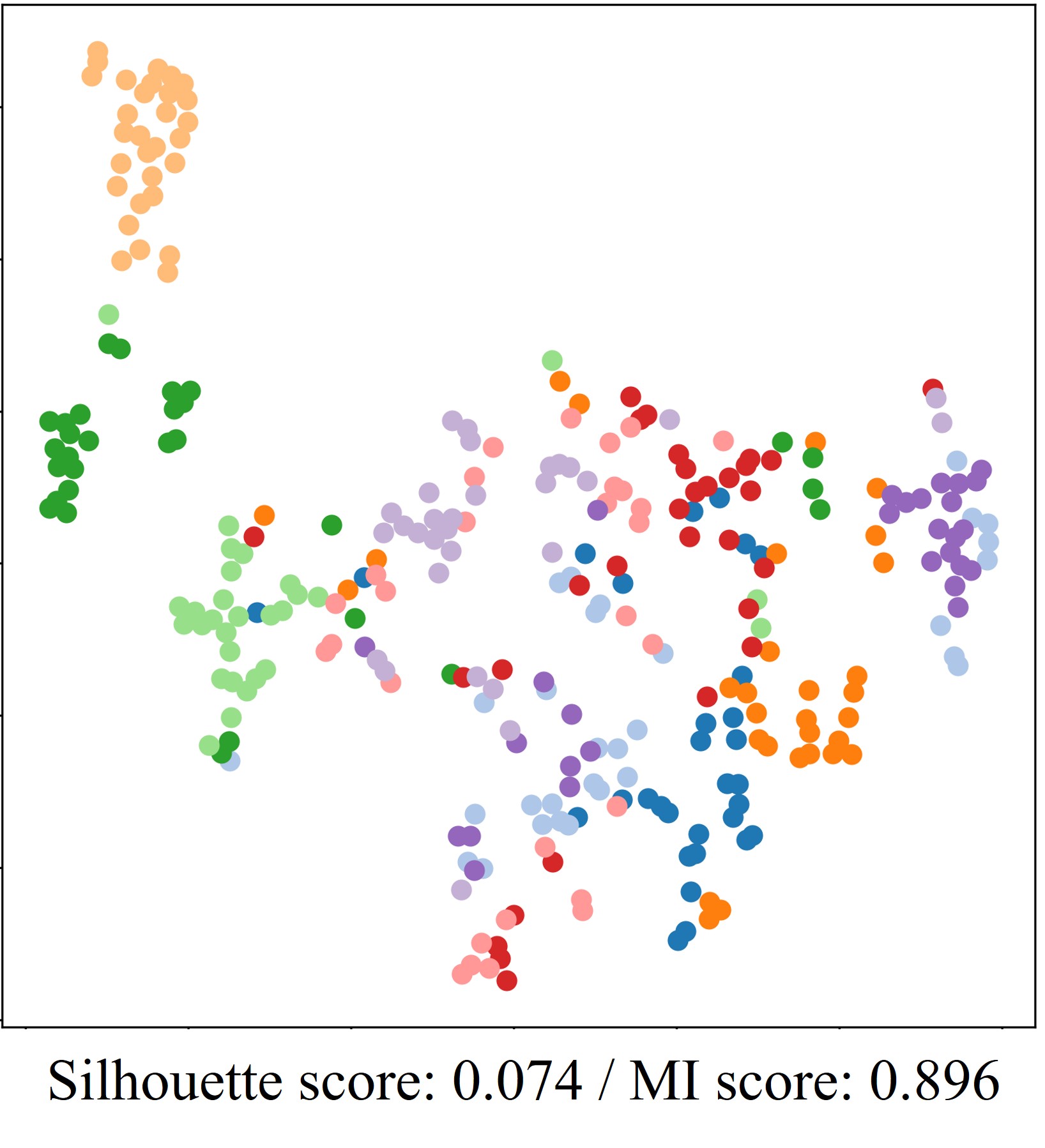}
    \caption{FRoST [ECCV'22]}
    \label{fig:viz_frost}
\end{subfigure}
\hfill
\begin{subfigure}{0.32\columnwidth}
    \centering
    \includegraphics[width=\columnwidth]{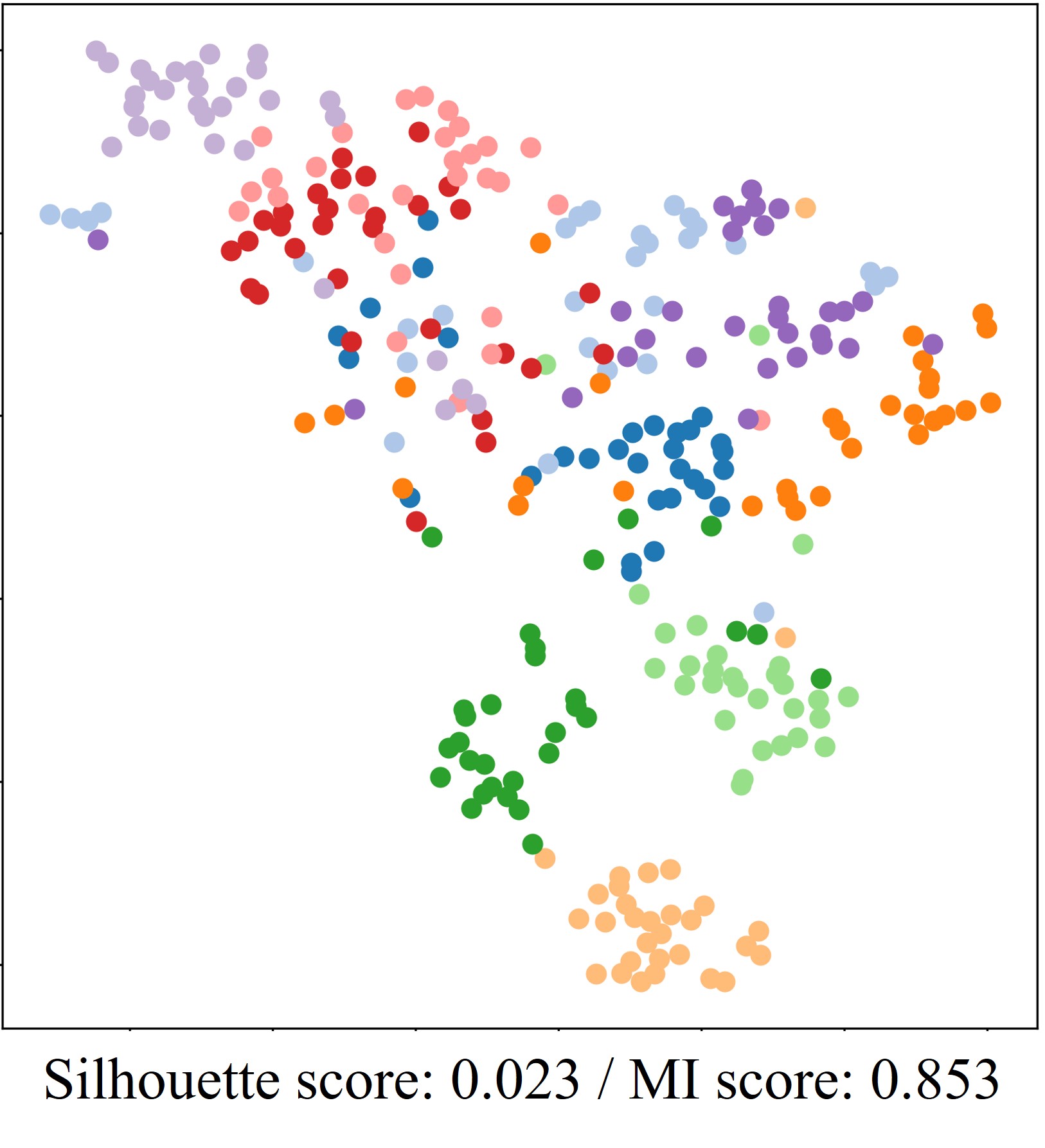}
    \caption{CGCD [ICCV'23]}
    \label{fig:viz_cgcd}
\end{subfigure}
\hfill
\begin{subfigure}{0.328\columnwidth}
    \centering
    \includegraphics[width=\columnwidth]{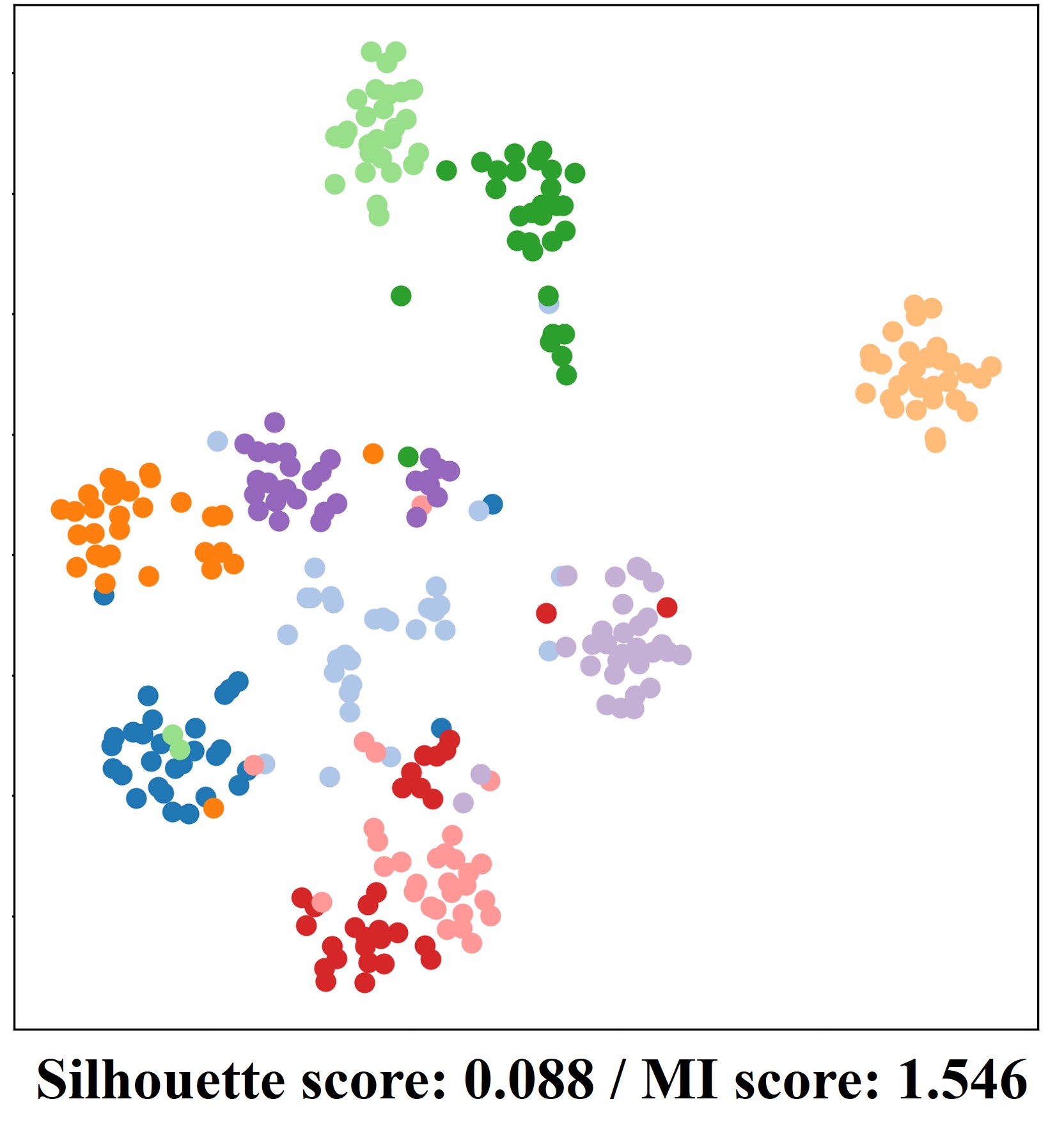}
    \caption{\textbf{DEAN (Ours)}}
    \label{fig:viz_dean}
\end{subfigure}
\caption{Comparison feature visualization for the novel categories on CUB200. We visualized 10 categories sequentially selected among the 40 novel categories.}
\label{fig:viz_all}
\end{figure}

\section{Visualization to Validate Novel Category Discovery}
We compared feature visualizations among FRoST, CGCD, and the proposed DEAN to validate the clustering performance on CUB200. Additionally, we measured the silhouette score and mutual information-based score to compare the effectiveness of clustering. The silhouette score indicates the quality of clustering, with a higher score indicating better clustering. A mutual information-based score measures the agreement between clustering labels and ground truth labels. a higher score indicates more effective labeling through clustering. As shown in~\Cref{fig:viz_all}, FRoST and CGCD faced challenges in clustering novel categories due to online continuous learning. Their performance metrics were notably lower compared to DEAN. For instance, DEAN achieved a mutual information-based score of 1.546, outperforming FRoST and CGCD significantly. This result highlights the ability of DEAN to effectively cluster novel categories and assign accurate labels compared to prior methods.

\section{Further Analysis of Energy-based Contrastive Loss}
\begin{figure}[t]
\begin{subfigure}{0.45\columnwidth}
    \centering
    \includegraphics[width=\columnwidth]{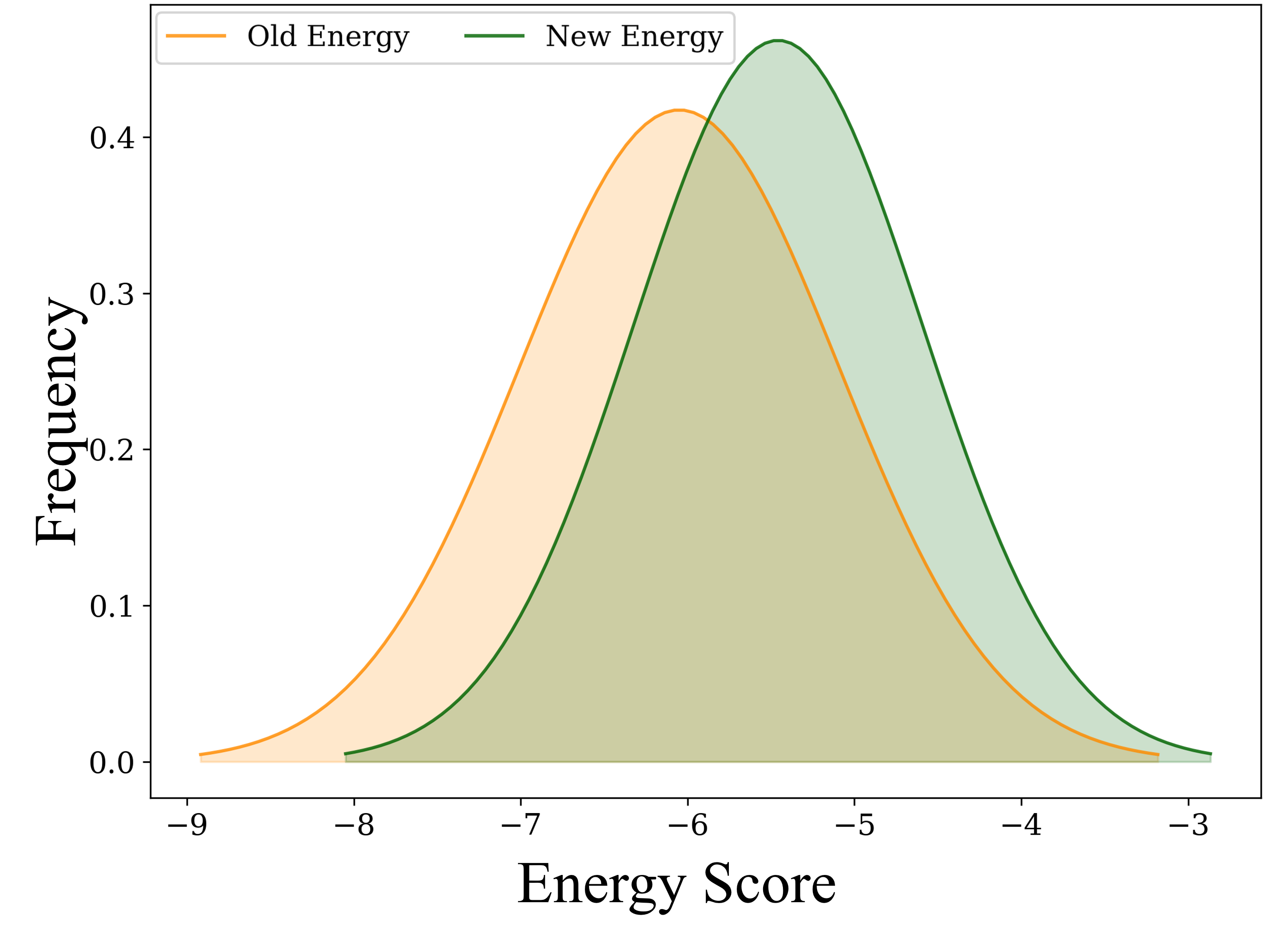}
    \caption{w/o EC Loss}
    \label{fig:viz_cgcd}
\end{subfigure}
\hfill
\begin{subfigure}{0.45\columnwidth}
    \centering
    \includegraphics[width=\columnwidth]{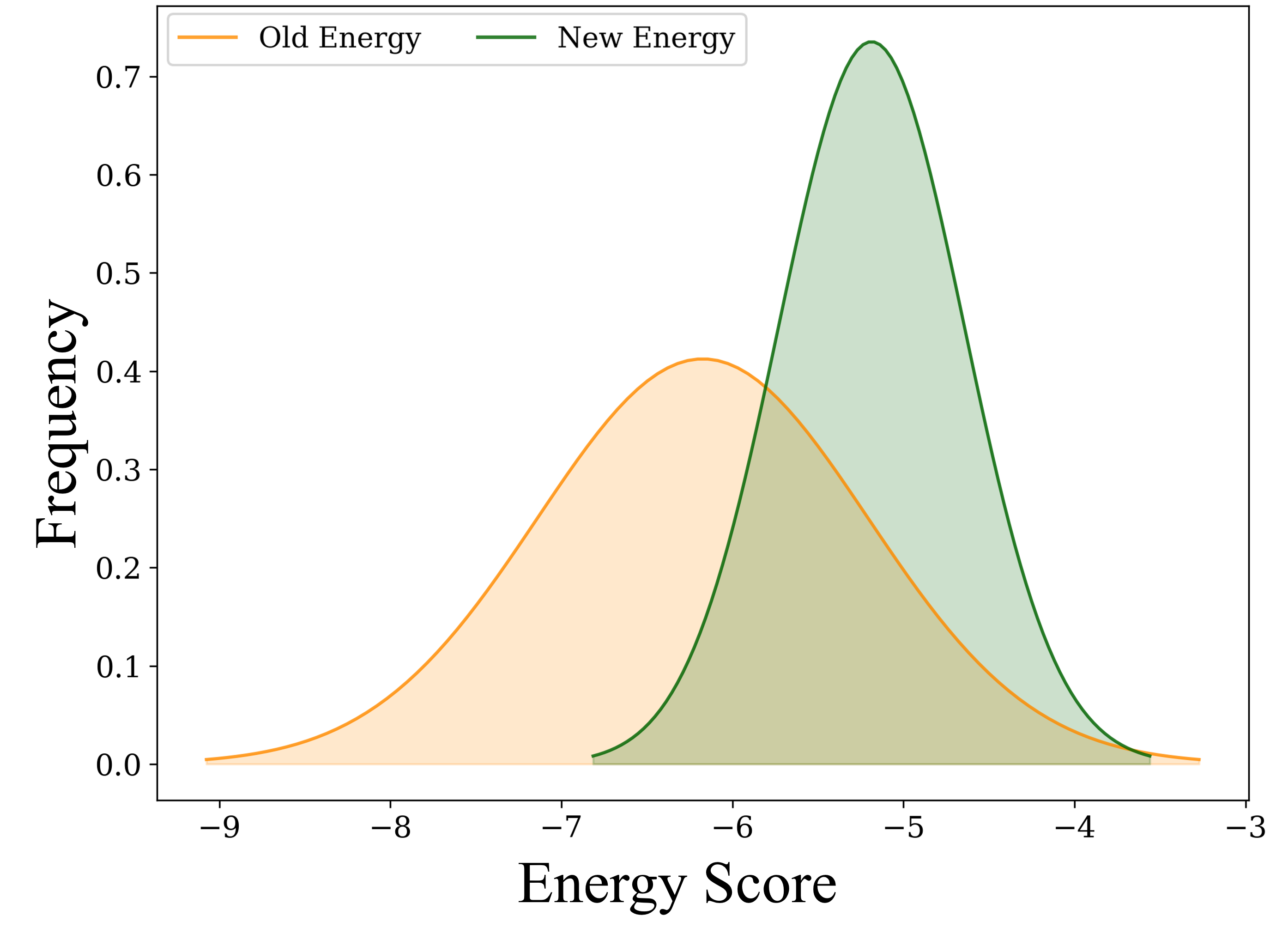}
    \caption{w/ EC Loss}
    \label{fig:enter-label}
\end{subfigure}
\caption{Comparing the normal distributions of energy scores using data estimated as unknown during the incremental session. Old energy and new energy denote the energy scores calculated by the known and unknown nodes from the online head, respectively.}
\label{fig:ec_loss_viz}
\end{figure}

To validate the effectiveness of the proposed energy-based contrastive loss, we conducted an ablation study using the energy scores of data estimated as unknown during the incremental session. Specifically, we calculated two different energy scores: old energy, based on the known nodes, and new energy, based on the unknown nodes from the online head. As shown in Figure \ref{fig:ec_loss_viz}, the application of the energy-based contrastive loss resulted in a more discernible difference between old energy and new energy. This experimental result demonstrates that the proposed energy-based contrastive loss assists the model in detecting data from novel categories, thereby facilitating the learning of novel categories.
\begin{table}[t]
\centering
\begin{minipage}{\columnwidth}
\caption{Ablation study of diverse variance values for variational feature augmentation. $\sigma_l$, $\sigma_{B^t}$, and $\sigma_u$ denote the standard deviation of feature vectors from the entire labeled dataset, provided batch-wise data, and detected unseen data, respectively.}
\label{tab:vfa_var}
\resizebox{\columnwidth}{!}{%
\setlength{\tabcolsep}{8.pt}
\renewcommand{\arraystretch}{1.3}
\begin{tabular}{cccccccc}
\toprule
\multirow{2}{*}{\large Variance $\sigma^2$ } & \multicolumn{7}{c}{\large CUB200} \\ \cline{2-8}
 & $M_{all}$ & $M_{old}$ & $M_{new}$ & $F\downarrow$ & $M^{PS}_{all}$ & $M^{PS}_{old}$ & $M^{PS}_{new}$ \\ \hline
w/o VFA & 57.37 & 64.98 & 27.63 & 17.61 & 57.31 & 85.19 & 46.47 \\ \hline 
$\hat{h}_u^t\sim\mathcal{N}(h_u, \sigma_l^2)$ & 55.56 & 65.44 & 26.86 & 17.35 & 75.72 & 93.39 & 64.41 \\ 
$\hat{h}_u^t\sim\mathcal{N}(h_u, \sigma_{B^t}^2)$ & 58.96 & 66.20 & 27.92 & 16.60  & 75.55 & 93.39 & 64.48 \\ \hline
\boldmath ${\hat{h}_u^t\sim\mathcal{N}(h_u, \sigma_u^2)}$ & \textbf{59.56} & \textbf{66.33} & \textbf{33.11} & \textbf{16.26} & \textbf{76.93} & \textbf{94.05} & \textbf{65.75} \\
\bottomrule
\end{tabular}%
}
\end{minipage}
\end{table}

\section{Further Analysis of Variational Feature Augmentation}
We introduced variational feature augmentation to enhance pseudo-labeling via accurate clustering. To analyze its effectiveness, we conducted an ablation study with diverse variance values on CUB200. Variance values were calculated from feature vectors of the entire labeled dataset, the batch-wise data provided during online learning, and the data detected as unknown. As shown in~\Cref{tab:vfa_var}, adopting variance values estimated from unknown data recorded superior clustering performance compared to other variance values. This experimental result demonstrates that leveraging variance information from estimated unknown data helps the model discover novel categories and improve pseudo-labeling accuracy.
Moreover, We conducted a comparison study between our VFA and AutoAug~\cite{Cubuk_2019_CVPR}, TrivialAug~\cite{muller2021trivialaugment}, and Moment Exchange (ME)~\cite{li2021feature} which do not require label information as VFA does. Table~\ref{tab:re_vfa} shows the promising performance of VFA, while the others faced significant shifts in feature space due to unstable online learning, hindering effective clustering.
\begin{table}[t]
\centering
\begin{minipage}{0.8\columnwidth}
\captionof{table}{Further experiment to evaluate the proposed online continuous generalized category discovery in various settings.}
\resizebox{\columnwidth}{!}{%
\setlength{\tabcolsep}{8.pt}
\renewcommand{\arraystretch}{1.}
\begin{tabular}{ccccc}
\specialrule{1}{1}{1}
\multirow{2}{*}{Augmentations} & \multicolumn{4}{c}{CUB200} \\ \cline{2-5}
 & $M_{all}$ & $M_{old}$ & $M_{new}$ & $F\downarrow$  \\ \hline
AutoAug {[}CVPR’ 19{]} & 57.90 & 65.76 & 27.20 & 16.28 \\
TrivialAug {[}ICCV '21{]} & 56.98 & 65.51 & 23.64 & 16.81 \\
ME {[}CVPR’ 21{]} & 57.73 & 66.27 & 24.35 & 19.02 \\
\hline
\textbf{VFA (Ours)} & \textbf{59.56} & \textbf{66.33} & \textbf{33.11} & \textbf{16.26} \\
\specialrule{1}{1}{1}
\end{tabular}%
}
\caption{Comparison of VFA against existing augmentations.}
\label{tab:re_vfa}
\end{minipage}
\end{table}

\bibliographystyle{splncs04}
\bibliography{main}